\documentclass[smallextended]{svjour3}

\usepackage{algorithm}
\usepackage{algorithmic}

\usepackage{graphicx}
\usepackage{bm}
\usepackage{natbib}

\begin{document}

\journalname{ArXiv version}

\title{HIVE-COTE 2.0: a new meta ensemble for time series classification}

\titlerunning{HIVE-COTE 2.0}

\author{Matthew Middlehurst \and James Large \and Michael Flynn \and Jason Lines \and Aaron Bostrom \and Anthony Bagnall}

\authorrunning{Middlehurst, et al.}

\institute{
    School of Computing Sciences, University of East Anglia, Norwich, UK \\
    Matthew Middlehurst, m.middlehurst@uea.ac.uk, https://orcid.org/0000-0002-3293-8779 \\
    James Large, james.large@uea.ac.uk, https://orcid.org/0000-0002-2357-3798 \\
    Michael Flynn, michael.flynn@uea.ac.uk, https://orcid.org/0000-0002-6811-5395 \\
    Jason Lines, j.lines@uea.ac.uk, https://orcid.org/0000-0002-1496-5941 \\
    Aaron Bostrom, a.bostrom@uea.ac.uk, https://orcid.org/0000-0002-7300-6038 \\
    Anthony Bagnall, ajb@uea.ac.uk, https://orcid.org/0000-0003-2360-8994 \\
}

\maketitle

\begin{abstract}
    The Hierarchical Vote Collective of Transformation-based Ensembles (HIVE-COTE) is a heterogeneous meta ensemble for time series classification. HIVE-COTE forms its ensemble from classifiers of multiple domains, including phase-independent shapelets, bag-of-words based dictionaries and phase-dependent intervals. Since it was first proposed in 2016, the algorithm has remained state of the art for accuracy on the UCR time series classification archive. Over time it has been incrementally updated, culminating in its current state, HIVE-COTE 1.0. During this time a number of algorithms have been proposed which match the accuracy of HIVE-COTE. We propose comprehensive changes to the HIVE-COTE algorithm which significantly improve its accuracy and usability, presenting this upgrade as HIVE-COTE 2.0. We introduce two novel classifiers, the Temporal Dictionary Ensemble (TDE) and Diverse Representation Canonical Interval Forest (DrCIF), which replace existing ensemble members. Additionally, we introduce the Arsenal, an ensemble of ROCKET classifiers as a new HIVE-COTE 2.0 constituent. We demonstrate that HIVE-COTE 2.0 is significantly more accurate than the current state of the art on 112 univariate UCR archive datasets and 26 multivariate UEA archive datasets.
    %PUT BACK IF WE INCLUDE THESE: To avoid claims of over-fitting, we introduce a number of new univariate and multivariate datasets not a part of either archive and replicate the success of HIVE-COTE 2.0 on them.

    \keywords{Time series classification \and Multivariate time series \and Heterogeneous ensembles \and HIVE-COTE}
\end{abstract}

\section{Introduction}

\begin{figure}[htb]
    	\centering
        \includegraphics[width=\linewidth,trim={1cm 8cm 0cm 4cm},clip]{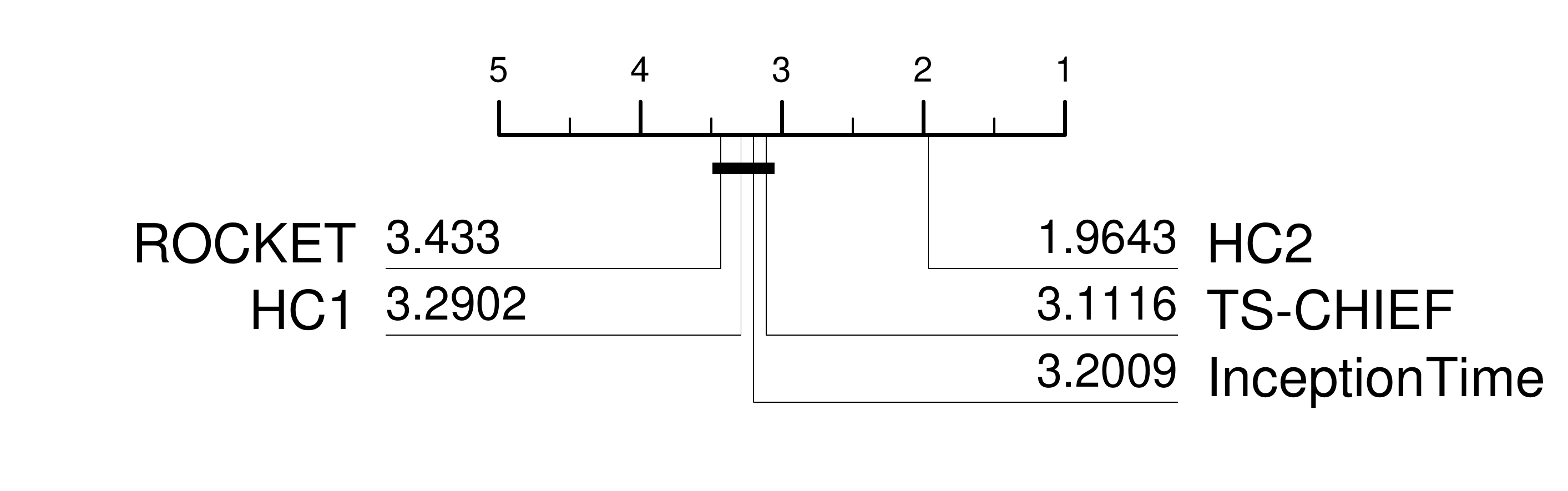}
        \caption{Critical difference diagram for HC2 against the current state of the art on 112 UCR TSC problems. The average rank for each classifier is shown, and solid lines group classifiers between which there is no significant difference. It demonstrates that there is no difference between HC1~\citep{bagnall20hivecote1}, InceptionTime~\citep{fawaz20inception}, ROCKET~\citep{dempster20rocket} and TS-CHIEF~\citep{shifaz20ts-chief}, but HC2 is significantly higher ranked than all of them. More details are given in Section~\ref{sec:results}.}
        \label{fig:acc-cd}
    \end{figure}

Time series classification (TSC) is the problem of predicting a discrete target variable from a (possibly multivariate) time series. TSC problems are seen in all areas of machine learning applications, including seizure detection~\citep{chaovalitwongse2006electroencephalogram}, earthquake monitoring~\citep{arul2021applications}, insect classification~\citep{potamitis2014classifying} and predictive maintenance~\citep{guillaume2020time}. The publication of the University of California, Riverside (UCR) TSC archive resulted in an increased interest into algorithmic research for this type of problem. An experimental study, characterised as a bake off~\citep{bagnall17bakeoff}, facilitated the objective and reproducible comparison of learning algorithm performance on the UCR archive. Since then, new classifiers have been proposed in the literature that have advanced the field by significantly outperforming those used in the bake off. There are currently four algorithms with reasonable claim to being state of the art for TSC based on experimentation on the recently expanded UCR archive~\citep{dau19ucr}. These are: the deep learning approach called InceptionTime~\citep{fawaz20inception}; the tree based Time Series Combination of Heterogeneous and Integrated Embedding Forest (TS-CHIEF)~\citep{shifaz20ts-chief}; the Random Convolutional Kernel Transform (ROCKET)~\citep{dempster20rocket}; and the heterogeneous meta-ensemble Hierarchical Vote Collective of Transformation-based Ensembles (HIVE-COTE)~\citep{lines18hive}, the latest version of which is called HIVE-COTE version 1.0 (HC1)~\citep{bagnall20hivecote1}. We propose a new version of HIVE-COTE that is significantly more accurate than all four current state-of-the-art algorithms. We call this
classifier HIVE-COTE version 2.0, or HC2 for short. The critical difference diagram in Figure~\ref{fig:acc-cd} summarises the final results of HC2 against the four leading algorithms on 112 equal length UCR archives, using 30 stratified resamples on each dataset (more detail is provided in Section~\ref{sec:results}). The number associated with each algorithm is the average rank of the classifier on 112 UCR datasets and solid bars group classifiers between which there is no significant difference. HC2 is on average over 1\% more accurate per problem than all of the current state of the art.

The key principle behind HIVE-COTE is that TSC problems are best approached by careful consideration of the data representation, and that with no expert knowledge to the contrary, the most accurate algorithm design is to ensemble classifiers built on different representations. The changes from HC1 to HC2 relate to the component classifiers and a redefinition of the underlying data representations used. HC2 contains four component classifiers: the dictionary based Temporal Dictionary Ensemble (TDE)~\citep{middlehurst20temporal}; the interval based Diverse Representation  Canonical Interval Forest (DrCIF)~\citep{middlehurst20canonical}; an adaptation of ROCKET we call the Arsenal; and the latest version of the Shapelet Transform Classifier (STC)~\citep{bostrom17binary}. Each of these classifiers represents the best in class for a particular representation. Prototype versions of TDE and DrCIF have been presented at conferences and are novel contributions in their own right. Arsenal enhances ROCKET to produce better probability estimates. STC has enhanced usability options that help improve the functionality of the whole ensemble; HC2 is now contractable (i.e. you can give the classifier a maximum run time), checkpointable (i.e. you can restart the classifier from a previous run) and works with multivariate time series classification (MTSC). A recent study~\citep{ruiz21mtsc} concluded that that MTSC is at an earlier stage of development than univariate TSC. The only algorithms significantly better than the standard TSC benchmark, one nearest neighbour with dynamic time warping (DTW), were HC1, ROCKET, InceptionTime and CIF~\citep{middlehurst20canonical}. HC2 is significantly more accurate than all these algorithms on the UEA MTSC archive~\citep{bagnall18mtsc}.

%We have conducted extensive experimentation using the UCR archive, and even with so many datasets, there is always a risk we have simply over fit the data to bias this results. To counter this, we evaluate HC2 on 24 completely new univariate TSC datasets. These are either padded versions of unequal length problems already in the archive (X) or data that have been donated to the archive since the 2018 batch update (X problems). We demonstrate that HC2 is also significantly more accurate on these data WE HOPE!!  The experimental set up is outlined in Section~\ref{sec:methodology} and the classifier comparisons are described in Section~\ref{sec:results}.

We investigate the effect of how HC estimates test error and the relative importance of each component of HC2 through an ablative study in Section~\ref{sec:ablative}. We show that there is high variability in performance between the components over datasets and that each component significantly improves the ensemble overall. We also assess alternative ensemble structures, including stacking and selection schemes, and conclude that the simple weighted structure using a tilted distribution, the Cross-validation Accuracy Weighted Probabilistic Ensemble (CAWPE)~\citep{large19cawpe}, is as good as much more complex approaches.

A common criticism of HC is its run time. This in part is due to out of date received wisdom relating to shapelet search. The initial shapelet search algorithms conducted a computationally expensive exhaustive search. This is not only unnecessary but results in over fitting. Recent versions of STC simply randomise the search. Nevertheless, a full run of HC2 is still computationally expensive on large problems and takes much longer than ROCKET. In Section~\ref{sec:usability} we explore the usability of HC2, including details of open source implementations and assess the performance and its components on two problems with very long series. Our results indicate that in general HC2 performance converges quickly, and that a contracted run time is sufficient to produce reasonable results in a controlled time. Finally, in Section~\ref{sec:conclusion} we conclude this study by identifying areas of future improvement for HIVE-COTE.

\section{Background}
\label{sec:background}

    %TSC/MTSC
    Time series classification (TSC) requires a training set of instance pairs $\{\bm{x},y\}$ with $m$ real-valued ordered observations and a discrete class label $y$ from a range of $c$ possible values. The objective is to create a function that maps from the space of possible input series to the space of possible class labels. This is achieved by using a training set of instance pairs to build a model that can output either a predicted class value, or a predicted class distribution, for previously unseen series.

     We restrict our attention to problems where series are the same length. In univariate TSC, $\bm{x}$ is a vector of $m$ observations.     The majority of research effort over the last decade has been into developing univariate TSC algorithms. Multivariate time series classification (MTSC) is an extension where the series are multidimensional and a single case is represented by a list of vectors over $d$ dimensions and $m$ observations, $\bm{X}=<\bm{x_1}, \bm{x_2}, ..., \bm{x_d}>$, and $\bm{x_k}=<x_{1,k}, x_{2,k}, ..., x_{m,k}>$. When indexing into a dataset, we denote the $j^{th}$ observation of the $i^{th}$ case in dimension $k$ as $x_{i,j,k}$.

One way of categorising algorithms is on the core data representation used. Distance based algorithms rely on elastic distance measures between two series.  Dictionary based approaches are based on the frequency of recurring patterns, found through converting real valued time series into a sequence of discrete symbol words. Interval based algorithms derive features on intervals of series to find temporal features that may be otherwise obscured by irrelevant observations. Shapelet based approaches find phase independent discriminatory subseries.

The current approaches to time series classification that exploit one or more of these representations can be grouped into four categories: modular heterogeneous ensembles where each module consists of a classifier built on a particular transformation type such as HIVE-COTE; tree based homogeneous ensembles where different data representations are embedded within the nodes of the tree~\cite{shifaz20ts-chief}; deep learning algorithms where the representations are embedded in the network~\citep{fawaz19deep}; and transformation/convolution approaches that create massive new feature spaces that are parsed with a linear classifier~\citep{dempster20rocket,nguyen19interpretable}. The best algorithms exploit one or more representations. In the remainder of this background, we limit ourselves to describing the most accurate approaches. More complete reviews of TSC algorithms can be found in~\citet{bagnall17bakeoff,bagnall20hivecote1}.

\textbf{HIVE-COTE 1.0.} The original version of HIVE-COTE was first introduced in 2016~\citep{lines16hive,lines18hive} and, at the time, was significantly more accurate on average than other known approaches~\citep{bagnall17bakeoff} on the 85 datasets that were then the complete UCR archive~\citep{dau19ucr}. The first version of HIVE-COTE (later dubbed HIVE-COTE alpha), contained five constituent ensembles that each worked on features from different data transformation domains: the Elastic Ensemble (EE)~\citep{lines15elastic}; Shapelet Transform Classifier (STC)~\citep{hills14shapelet}; Time Series Forest (TSF)~\citep{deng13forest}; Bag of Symbolic-Fourier-Approximation Symbols (BOSS)~\citep{schafer15boss}; and the Random Interval Spectral Ensemble (RISE) that was introduced alongside HIVE-COTE~\citep{lines18hive}. Each module was encapsulated and built on the train data independently of the others. For new data, each module passes an estimate of class probabilities to the control unit, which combines them to form a single prediction. It does this by weighting the probabilities of each module by an estimate of its testing accuracy formed from the training data.

    The goal of HIVE-COTE alpha was to achieve the highest level of accuracy without concern for computational resources. This initial target has since lead to a perception that HIVE-COTE is very slow and does not scale well. A very simple restructure of HIVE-COTE alpha was able to achieve the same level of accuracy in orders of magnitude less time; HIVE-COTE 1.0 (HC1)~\citep{bagnall20hivecote1}, was introduced to demonstrate its utility and scalability. (HC1) is based on simple refinements and enchantments to the original HIVE-COTE alpha base constituents. HC1 dropped the distance based EE due to the high computational overhead. STC introduced binary shapelets and a randomised search controlled by a time parameter. HC1 uses the Cross-validation Accuracy Weighted Probabilistic Ensemble (CAWPE)~\citep{large19cawpe} ensemble structure. CAWPE uses an accuracy estimate of each classifier formed on the train data to weight the probabilities of each component. It constructs a tilted distribution through exponentiation using a parameter $\alpha$ to extenuate differences in classifiers. Each component's weight is found through an internal estimate for each classifier if capable, else a ten fold cross-validation of the training data is performed.

    %An overview of the HIVE-COTE 1.0 structure is displayed in Figure~\ref{fig:hc1.0}.
%MAYBE PUT THE HC1 PICTURE BACK?

\textbf{TS-CHIEF.}
   The Time Series Combination of Heterogeneous and Integrated Embedding Forest (TS-CHIEF)~\citep{shifaz20ts-chief} is the classifier most comparable to HIVE-COTE. TS-CHIEF is made up of an ensemble of trees which embed distance, dictionary and spectral base features. A number of splitting criteria from each representation with randomly initialised parameters are considered at each node. The different types of split criteria are dictionary based splits based on BOSS, similarity based splits based on EE and interval based splits based on RISE. The core distinction is that the usage of base features is embedded in nodes of the tree rather than modularised through separate classifiers.

\textbf{InceptionTime.}~\citep{fawaz20inception} is a deep learning ensemble, combining five homogeneous residual networks incorporating inception modules~\citep{szegedy15inception}. An individual network is made up of two blocks of three Inception modules which maintain residual connections, followed by global average pooling and softmax layers. Each network in the ensemble is initialised with random weights for stability. It is the best deep learning approach for time series data to our knowledge, and represents deep learning for TSC in our experiments.

\textbf{ROCKET.} the Random Convolutional Kernel Transform (ROCKET)~\citep{dempster20rocket} produces a large number of summary stats using randomly initialised convolutional kernels, then selects informative ones using a linear classifier. A version of ROCKET is included in the HIVE-COTE 2.0 ensemble, so a more complete description of the algorithm is provided in Section \ref{sec:arsenal}.

Other recent approaches focus on a single representation. Proximity Forest~\citep{lucas19proximity} is a tree ensemble that randomly chooses distance functions at each node. Supervised Time Series Forest (STSF)~\citep{cabello20fast} is an interval based tree ensemble that includes a supervised method for extracting intervals and uses summary statistics and spectral features.
    A number of extensions to the BOSS classifier have been made since the bakeoff in S-BOSS~\citep{large19dictionary}, cBOSS~\citep{middlehurst19scalable} and WEASEL~\citep{schafer17fast}.

There have also been a range of algorithms proposed for MTSC~\citep{ruiz21mtsc}. Dynamic Time Warping with pointwise multivariate distance and a one nearest neighbour classifier, characterised as dependent dynamic time warping (DTW-D)~\citep{shokoohi2017generalizing}, is the baseline for MTSC. ROCKET, InceptionTime and CIF have multivariate versions which are significantly more accurate.

\section{HIVE-COTE 2.0 (HC2)}
\label{sec:hc2}

\begin{figure}[htb]
    	\centering
        \includegraphics[width=\linewidth,trim={3cm 1cm 3cm 1cm},clip]{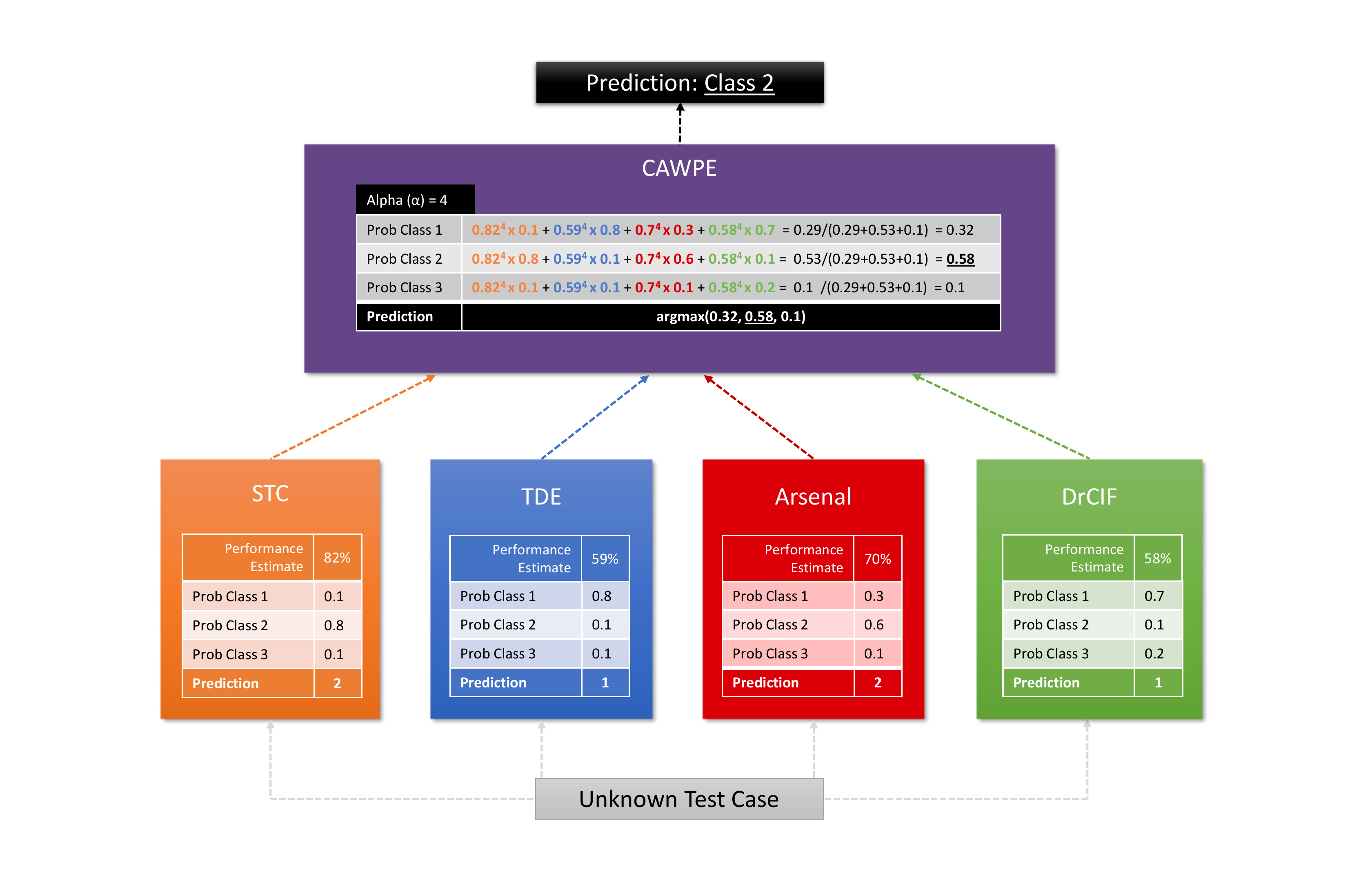}
        \caption{An overview of the ensemble structure of HIVE-COTE 2.0 for a three class problem. Each module is trained independently and produces an estimate of the probability of membership of each class for unseen data. The control unit (CAWPE) combines these probabilities, weighted by an estimate of the quality of the module found on the train data.}
        \label{fig:hc2.0}
    \end{figure}

HIVE-COTE 2.0 replaces three of the four classifiers that make up HIVE-COTE 1.0. The component modules are: the shapelet based Shapelet Transform Classifier~\citep{bostrom17binary}; the convolution based ensemble of ROCKET classifiers we call the Arsenal; the dictionary based representation TDE; and the interval based DrCIF. An overview of the updated HC2 structure is displayed in figure~\ref{fig:hc2.0}.
Each component is trained independently and in addition to the final model, it is required to produce an estimate of its accuracy on unseen data. For new data, each module produces a probability estimate for each class. The controller constructs a tilted distribution through exponentiation (using $\alpha = 4$ by default) to extenuate differences in classifiers and weighting with the accuracy estimate. Each module of HC2 contains new features and improvements over previous versions. These include novel algorithm improvements, multivariate extensions and contracting improvements. In addition, the method for estimating the accuracy from the train data has been improved. Generally, there are three ways of estimating test accuracy from train data. Firstly, you can assess the final model directly on the train data. This is likely to be biased and over optimistic, particularly if some form of model selection has occurred without careful regularisation. Secondly, you can cross validate on the train data in addition to the final build. Whilst this will probably be less biased (and generally pessimistic) it is time consuming. Thirdly, you can use some form of hold-out evaluation embedded in the full model build, such as bagging. HC1 uses a mixture of approaches based on the classifier. HC2 adopts a standardised bagging approach, which is possible since all four classifiers in HC2 use ensembles. However, we found that whilst using out of bag performance produces acceptable estimates of test accuracy, the bagged classifiers themselves were significantly less accurate than those built on the full data. Hence, we adopted a hybrid approach. Rather than build 11 total models for ten fold cross validation or a single model using bagging, we construct one bagged model to estimate accuracy for those that require it, and a full model to predict new cases. The impact of this design choice is discussed in Section~\ref{sec:ablative}. %TDE uses oob, but only has a single model as it naturally subsamples 70% of train data

HC2 can train each component concurrently. Even so, the components can be slow on large problems. Hence, we allow the user to configure HC2 so that it has a time contract. If contracted, each component ensemble simply builds as many base classifiers as it can in the time provided. This simple form on contracting is an adequate first fix. However, a problem does arise for very large data with short contracts: building a single ensemble member may exceed the contract. It would be better for the components to self configure when this is likely, by, for example, subsampling cases or series. HC2 is threaded but currently, the components themselves are not. Given they are all ensembles of independent base classifiers, it is in principle easy to do so. This and the inevitable move onto GPU is part of our future work plan.

    \subsection{Temporal Dictionary Ensemble (TDE)}
    \label{sec:tde}

    HIVE-COTE alpha contains the dictionary based classifier BOSS~\citep{schafer15boss}, which was updated to cBOSS~\citep{middlehurst19scalable} in HC1. HC2 uses the Temporal Dictionary Ensemble (TDE) (first introduced in~\citet{middlehurst20temporal}), which draws on more recent work on dictionary classifiers~\citep{large19dictionary,schafer17fast} and includes several novel features. Dictionary based approaches aim to capture the repetitions of patterns as discriminatory features rather than solely their presence. These approaches commonly adapt the bag-of-words model used in other domains such as signal processing, computer vision and audio processing for time series data.

    TDE is an ensemble of 1-NN classifiers that transforms each series into a histogram of word counts. A sliding window of length $w$ is run along each series, and the subseries is discretised into a word of length $l$ from an alphabet of size $\alpha$. TDE transforms the window using the Symbolic Fourier Approximation (SFA)~\citep{schafer12sfa} transform proposed for BOSS~\citep{schafer15boss}. Distance between histograms is found using histogram intersection. In addition to word frequencies, TDE also captures the frequencies of bigrams found from non overlapping windows. Thus a transformed instance includes a histogram of word counts and bigram counts for a given trio of parameters ($w$,$l$,$\alpha$). TDE also includes some spatial information by the utilisation of spatial pyramids~\citep{lazebnik06pyramid}. This involves splitting a series into $h$ levels each with $2^{v}$ disjoint subseries, where $v$ is the current pyramid level. Word counts are found for each subseries independently, then the resulting histograms are concatenated.   The distance to histograms of deeper levels with smaller spatial areas in the series are weighted higher than global similarity. Bigrams are only recorded for the first level consisting of the full series. The SFA transform requires a set of breakpoints when creating words. The method of generating these breakpoints $b$ is selected between Multiple Coefficient Binning (MCB)~\citep{schafer15boss} and Information Gain Binning (IGB)~\citep{schafer17fast}. Windows can optionally be normalised during the transform with the $p$ parameter.

    The TDE ensemble is filtered into $s$ total classifiers from $k$ candidates. The accuracy of each candidate is estimated using  leave-one-out cross-validation (LOOCV), with the highest $s$ being retained. Diversity is achieved through altering the parameters ($w$,$l$,$h$,$b$,$p$) for each new classifier and a 70\% sampling of the train data. The first 50 classifiers use randomly selected parameters, while those after are selected using a Gaussian processes regressor. For unseen parameter sets, a prediction of accuracy is made using the parameters of previously built classifiers, with the highest predicted accuracy being chosen for the next classifier build. New cases are classified with a weighted majority vote, using the exponential accuracy weights from CAWPE~\citep{large19cawpe}.
    Table~\ref{tab:TDEparameters} shows the range from which individual classifier parameters are subsampled from, with $m$ being the series length.
    The ensemble build process for TDE is described in Algorithm~\ref{alg:TDE}.
    When it comes to replacing BOSS in the HIVE-COTE ensemble, TDE was the only dictionary based approach to significantly improve the ensemble's accuracy~\citep{middlehurst20temporal}. Using cBOSS and S-BOSS as replacements was found to show no significant difference in accuracy, while WEASEL was significantly worse.

    We introduce the capability for multivariate time series classification to TDE by making a number of additions to the individual classifier. WEASEL also has a multivariate version, WEASEL-MUSE~\citep{schafer17multivariate}, which shares similarities with our extension. Current dictionary approaches are noticeably memory intensive due to the requirement to store multiple transformed versions of the original data. WEASEL-MUSE can become unusable due to this issue, with over 500GB of memory required on some datasets~\citep{ruiz21mtsc}. We aim to mitigate this issue with the TDE multivariate extension.

    For each dimension we extract words using the same process for univariate series, with each dimension having their own set of breakpoints. Words from different dimensions are stored separately in each instances bag. For many multivariate time series, some dimensions hold little or redundant information. Additionally, for problems with many dimensions storing the words extracted from each can cause significant memory issues. As such, prior to creating any bags we take a subsample of dimensions based on an accuracy estimate. We find this estimate using LOOCV on bags created from disjoint windows rather an a sliding one. Any dimension with an accuracy estimate less than 85\% of the highest accuracy is not retained for this classifier. Additionally, we set a limit of 20 maximum dimensions retained for each classifier, keeping those with the highest accuracy. To reduce the memory impact of saving features from multiple dimensions, we do not record bigrams for multivariate datasets. The build process for the individual classifiers used in TDE is displayed in Algorithm~\ref{alg:baseTDE}.

\begin{figure}[!htb]
    	\centering
        \includegraphics[width=\linewidth,trim={1cm 8cm 2cm 4cm},clip]{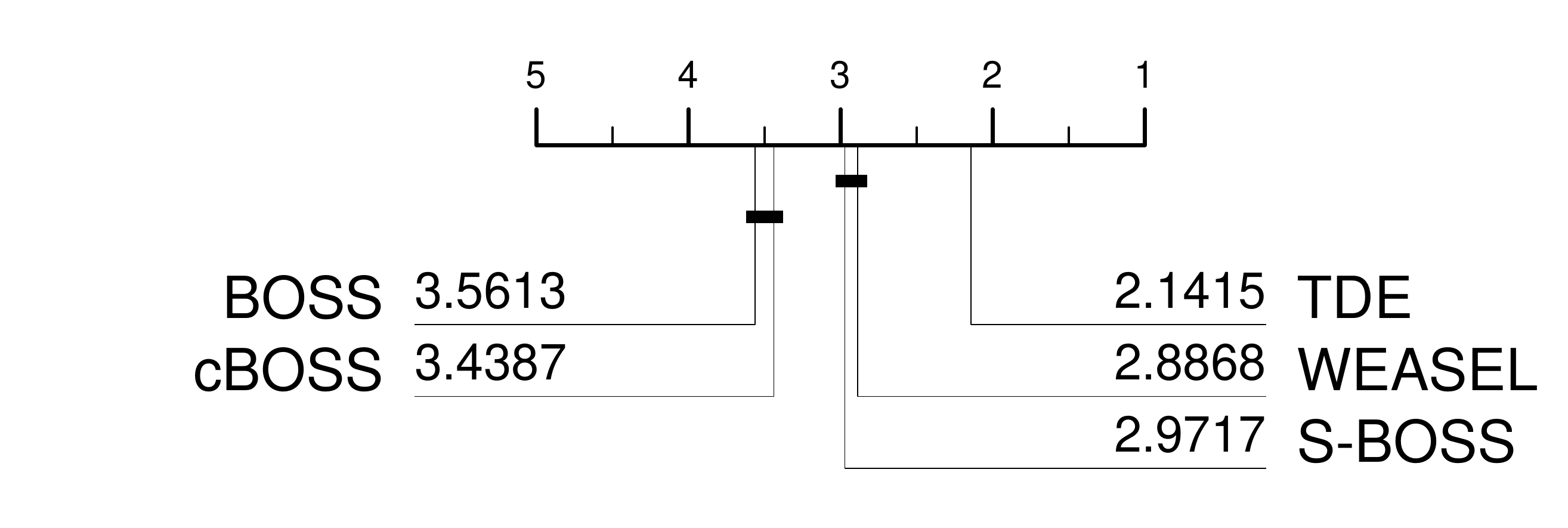}
        \caption{Results of five dictionary based classifiers on 106 of the UCR datasets. The missing datasets are: ElectricDevices; FordA; FordB; HandOutlines; Non-InvasiveFetalECGThorax1; and NonInvasiveFetalECGThorax2. These are missing due to the long run time of S-BOSS and WEASEL. cBOSS samples 250 parameter sets and has an ensemble size of 50. WEASEL $\chi$ is set to 2.}
        \label{fig:dictionaryCD}
    \end{figure}

    Figure~\ref{fig:dictionaryCD} provides evidence to support our claim that TDE is the most accurate purely dictionary based algorithm. It shows the results of a comparison of dictionary based classifiers on test accuracy performance on the UCR datasets. TDE is significantly more accurate than WEASEL and S-BOSS, which in turn are more accurate than BOSS and cBOSS.

    \begin{algorithm}[!htb]
        \caption{Temporal Dictionary Ensemble(A list of $n$ cases of length $m$ with $d$ dimensions, ${\bf T}=({\bf X,y})$)}
        \label{alg:TDE}
        \begin{algorithmic}[1]
            \REQUIRE the number of parameter samples $k$, the max ensemble size $s$ (default $k=250$, $s=50$)
            \STATE Let $w$ be window length, $l$ be word length, $p$ be normalise/not normalise, $\alpha$ be alphabet size, $h$ be number of pyramid levels and $b$ be MCB or IGB discretisation.
            \STATE Let ${\bf C}$ be a list of $s$ classifiers $({\bf c}_1,\ldots,{\bf c}_s)$
        	\STATE Let ${\bf E}$ be a list of $s$ classifier weights $({\bf e}_1,\ldots,{\bf e}_s)$
        	\STATE Let ${\bf G}$ be a list of $k$ parameter and accuracy pairs $({\bf g}_1,\ldots,{\bf g}_k)$
        	\STATE Let ${\bf R}$ be a set of possible parameter combinations
        	\STATE $i \leftarrow 0$
        	\STATE $lowest\_acc \leftarrow \infty, lowest\_acc\_idx \leftarrow \infty$
        	\WHILE {$i < k$ AND $|{\bf R}| > 0$}
                \IF{$i < 50$}
        		    \STATE $[l,\alpha,w,p,h,b] \leftarrow$ randomSample(${\bf R}$)
        		\ELSE
        		    \STATE $gp \leftarrow$ buildGaussianProcesses(${\bf G}$)
        		    \STATE $[l,\alpha,w,p,h,b] \leftarrow$ bestPredictedParameters(${\bf R}, gp$)
        		\ENDIF
        		\STATE $ {\bf R} = {\bf R} \setminus\{[l,\alpha,w,p,h,b]\} $
        		\STATE ${\bf T'} \leftarrow$ subsampleData(${\bf T}$, 0.7)
        		\STATE $cls \leftarrow$ buildIndividual(${\bf T'},l,\alpha,w,p,h,b$)
        		\STATE $acc \leftarrow$ LOOCV($cls$) \COMMENT{ {\em train data accuracy}}
        		\IF{$i < s$}
        		    \IF{$acc < lowestAcc$}
        		        \STATE $lowestAcc \leftarrow acc$, $lowestAccIndex \leftarrow i$
        	    \ENDIF
        		    \STATE $c_i \leftarrow cls$, $e_i \leftarrow acc^4$
        		\ELSIF{$acc > lowestAcc$}
        		    \STATE $c_{lowestAccIndex} \leftarrow cls$, $e_{lowestAccIndex} \leftarrow acc^4$
        		    \STATE $[lowestAcc,lowestAccIndex] \leftarrow$ findNewLowestAcc(${\bf C}$)
        		\ENDIF
        	    \STATE $g_i \leftarrow \{[l,\alpha,w,p,h,b],acc\}$
        		\STATE $i \leftarrow i+1$
        	\ENDWHILE
        \end{algorithmic}
    \end{algorithm}

    \begin{algorithm}[]
    	\caption{Individual TDE(A list of $n$ cases of length $m$ with $d$ dimensions, ${\bf T}=({\bf X,y})$)}
    	\label{alg:baseTDE}
    	\begin{algorithmic}[1]
        \REQUIRE word length $l$, alphabet size $\alpha$, window length $w$, normalisation parameter $p$, No. pyramid levels $h$, breakpoint function $b$
    		\STATE Let ${\bf H}$ be a list of $n$ histograms $({\bf h}_1,\ldots,{\bf h}_n)$
    		\STATE Let ${\bf K}$ be a list of dimension indexes found using LOOCV
    		\STATE Let ${\bf B}$ be a list of $|{\bf K}|$ $l$ by $\alpha$ matrices of breakpoints found by $b$ $({\bf b}_1,\ldots,{\bf b}_{|{\bf K}|})$
    		\STATE Let ${\bf O}$ be a matrix of $|{\bf K}|$ by $n$ by $m-w+1$ containing the words for each window
    		\FOR {$i \leftarrow  1$ to $n$}
    		    \FOR {$g \leftarrow 1$ to $|{\bf K}|$}
    			    \FOR {$j \leftarrow 1$ to $m-w+1$}
        				\STATE ${\bf s}\leftarrow x_{i,{k_g},j} \ldots x_{i,{k_g},j+w-1}$
        				\STATE ${\bf q} \leftarrow$ DFT(${\bf s}, l, \alpha$) \COMMENT{ {\em {\bf q} is a vector of the complex DFT coefficients}}
        				\IF{$p$}
            			    \STATE ${\bf q'} \leftarrow (q_2 \ldots q_{l/2+1})$
            			\ELSE
            			    \STATE ${\bf q'} \leftarrow (q_1 \ldots q_{l/2})$
            			\ENDIF
        				\STATE $r \leftarrow$ SFAlookup(${\bf q'}, b_{g}$)
        				\IF{$r \neq {o}_{g,i,j-1}$}
        				    \FOR {$v \leftarrow 1$ to $h$}
            					\STATE $pos \leftarrow $index(spatialPosition($r$,$v$),$g$)
            					\STATE ${h}_{i,pos} \leftarrow {h}_{i,pos} + 1$
            				\ENDFOR
        					\STATE $bi \leftarrow $index(bigram($r$,${o}_{g,i,j-w}$),$g$)
        					\STATE ${h}_{i,bi} \leftarrow {h}_{i,bi} + 1$
        				\ENDIF
        				\STATE ${o}_{g,i,j} \leftarrow r$
    				\ENDFOR
    			\ENDFOR
    		\ENDFOR
    	\end{algorithmic}
    \end{algorithm}

    \begin{table}[!htb]
        \centering
        \caption{Parameter ranges for TDE base classifier selection.}
        \begin{tabular}{l|l}
            \hline
            Parameter       & Range \\
            \hline
            Word lengths    & $l=\{16,14,12,10,8\}$\\
            Window lengths  & $w=\{10...m\}$  \\
            Normalise       & $p =\{true,false\}$ \\
            Alphabet Size & $\alpha = \{4\}$  \\
            No.pyramid levels  & $h =\{1,2,3\}$ \\
            Breakpoints        &  $b =\{MCB,IGB\}$ \\
            \hline
        \end{tabular}
        \label{tab:TDEparameters}
    \end{table}

    \subsection{Diverse Representation Canonical Interval Forest (DrCIF)}
    \label{sec:drcif}

    The Diverse Representation Canonical Interval Forest (DrCIF) is an interval based ensemble and an extension of its prototype version, the Canonical Interval Forest (CIF)~\citep{middlehurst20canonical}. Interval based classifiers extract phase-dependent subseries, aiming to find discriminatory features over different intervals. For time series of length $m$ there are $m(m-1)/2$ possible intervals that can be extracted. The original interval based classifier, the Time Series Forest~\citep{deng13forest}, is a component of HC1. It selects multiple intervals for each decision tree base classifier, then concatenates derived features (mean, standard deviation and slope) to form a diverse training set for each ensemble member. The other interval based classifier in HC1, RISE, selects a single interval for each base classifier, then derives spectral features (periodogram and auto-regressive terms) over the single interval. DrCIF replaces both these interval based classifiers, combining and enhancing both feature spaces. It draws on recent ideas presented in the STSF interval based classifier~\citep{cabello20fast} and the feature set method defined as the canonical time series characteristics (catch22)~\citep{lubba19catch22}. The catch22 features are a set of 22 features designed for time series data filtered from the 7658 features available in the highly comparative time series analysis (hctsa) toolbox~\citep{fulcher17hctsa}. After a pruning process, the catch22 features were derived from a clustering and filtering of the 7658 hctsa features based on accuracy, scalability and interpretability.

\begin{figure}[!htb]
    	\centering
        \includegraphics[width=\linewidth,trim={1cm 9cm 2cm 4cm},clip]{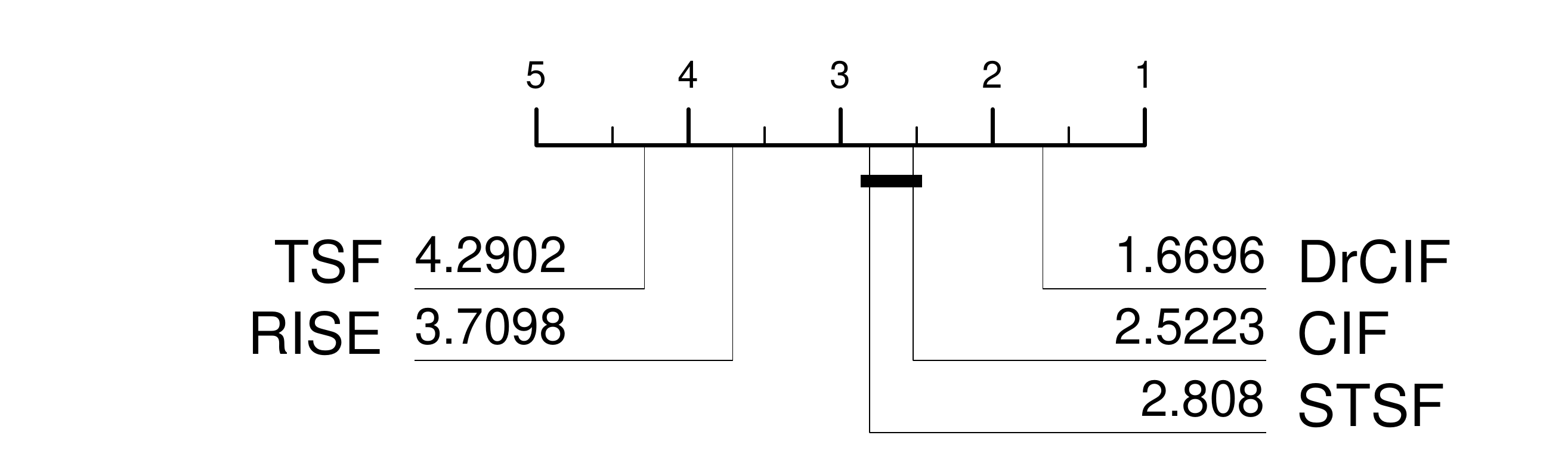}
        \caption{Critical difference diagram for five interval based classifiers on 112 UCR datasets. Each classifier builds 500 trees. TSF and CIF extract sqrt($m$) intervals per tree. CIF subsamples 8 attributes per tree.}
        \label{fig:intervalCD}
    \end{figure}

    The base classifier for DrCIF is a simple information gain based tree used in TSF, called the time series tree~\citep{deng13forest}. Features from the tree are derived from multiple intervals taken from the base series, the first order difference series and the periodograms of the whole series. Intervals from each are randomly selected.  Seven basic summary statistics are part of a pool of possible features extracted from an interval of any one of the three representations. These are: the mean; standard-deviation; slope; median; inter-quartile range; min; and max. DrCIF adds the catch22 features to this selection of summary statistics to form a candidate pool of 29 features.  $a$ out of the 29 features available are randomly selected for each tree. For each of the 3 representations, $k$ phase dependent intervals with randomly selected positions and lengths are extracted. The selected features are then calculated from each interval. These features are concatenated into a $3 \cdot k \cdot a$ length vector for each series, and the new dataset is used to build the tree. Diversity is achieved by providing each base classifier with different intervals and a different subset of the 29 features. Generally, we select $k$ as a function of the representation series length $rm$. Each representation will differ in its length, with the periodogram being half the size of the base series and the differences having one less value. As such it is likely the number of intervals selected for each representation will differ. For multivariate data, DrCIF randomly selects the dimension used for each interval. Replacing TSF with CIF in HIVE-COTE Alpha has been shown to significantly improve the classifier on univariate data~\citep{middlehurst20canonical}. The build process and the default parameter values
    for the DrCIF ensemble is described in Algorithm~\ref{alg:DrCIF}. Figure~\ref{fig:intervalCD} demonstrates that DrCIF represents a new best in class interval based classifier.

    \begin{algorithm}[!htb]
        \caption{Diverse Representation Canonical Interval Forest(A list of $n$ cases of length $m$ with $d$ dimensions, ${\bf T}=({\bf X,y})$)}
        \label{alg:DrCIF}
        \begin{algorithmic}[1]
            \REQUIRE the number of trees, $r$, the number of intervals per representation for each tree, $k$, and the number of attributes subsampled per tree, $a$ (default $r=500$, $k=4+(\sqrt{d}\sqrt{rm}$)/3, and $a=10$)  \COMMENT{ {\em where rm is the length of a representations series}}
            \STATE Let $\bf{F} = (F_1 \ldots F_r)$ be the trees in the forest
            \STATE Let $\bf{V}$ be a $3$ by $n$ by $d$ list of series with variable length, containing the base series list $\bf X$ and $\bf X$ transformed to periodograms and first order differences
        	\FOR {$i \leftarrow 1$ to $r$}
                \STATE Let $\bf{S}$ be a list of $n$ cases $(s_1 \ldots s_n)$ with $a \cdot k$ attributes
                \STATE Let $\bf{U}$ be a list of $a$ randomly selected attribute indices $(u_1 \ldots u_a)$
                \FOR {$y \leftarrow 1$ to $3$}
            		\FOR {$j \leftarrow 1$ to $k$}
            		    \STATE $b = rand(1,$$\bf |V_{y}|$$-3)$ \COMMENT{ {\em interval position}}
            		    \STATE $l = rand(3,$$\bf |V_{y}|$$/2)$ \COMMENT{ {\em interval length}}
            		    \STATE $o = rand(1,d)$ \COMMENT{ {\em interval dimension}}
            		    \FOR {$t \leftarrow 1$ to $n$}
            		        \FOR {$c \leftarrow 1$ to $a$}
            		            \STATE $s_{t,a(j-1)+c} \leftarrow summaryStat(u_c,$$\bf V_{y,t,o}$$,b,l)$
            		        \ENDFOR
            		    \ENDFOR
            		\ENDFOR
        		\ENDFOR
        		\STATE $F_i.buildTimeSeriesTree([$$\bf S$$,y])$
            \ENDFOR
        \end{algorithmic}
    \end{algorithm}

    \subsection{The Arsenal: A ROCKET Ensemble}
    \label{sec:arsenal}

    The Random Convolutional Kernel Transform (ROCKET)~\citep{dempster20rocket} uses a large number of randomly parameterised convolution kernels applied to each instance. As each kernel is applied to a series, the max value and proportion of positive values are recorded and concatenated into a feature vector. These features are then used to build a linear ridge regression classifier with built in cross-validation to select the alpha parameter.

    For each kernel generated, the parameters are selected from the following spaces: The length, $l$, is selected such that, $l \in \{7, 9, 11\}$; the value of each weight, $w_i$, is randomly sampled from a normal distribution $\sim \mathcal{N}(0,1)$, and are then mean centered; bias $b$ is sampled from a uniform distribution $\sim \mathcal{U}(-1,1)$; dilation, $a$, is sampled from an exponential scale up to series length; the binary decision to pad the series $p$ is chosen with equal probability, if true the series is zero padded at the start and end equally such that middle element of the kernel is applied to every point in the input series. Stride is always set to 1. For multivariate datasets, each kernel is assigned a random number of randomly selected dimensions. The kernel for the multivariate case is still one dimensional, but with weighting being different for each dimension. The max and proportion of positive values is calculated across all selected dimensions.

    ROCKET is a very fast classifier that has state-of-the-art accuracy, and we believe it is the most important recent development in the field. It represents a different class of approach, and as such is a candidate for assimilation into the collective. However, an issue arises when trying to include ROCKET in HIVE-COTE: the ridge regressor used by ROCKET is hard to configure to produce useful probability values for each class when making predictions. The CAWPE ensemble structure of HIVE-COTE uses weighted probabilities, and relies on classifiers to produce a distribution representative of the classifiers strength of belief in predictions. One solution would be to replace the ridge regressor with a classifier that does produce representative probability estimates. However, our experimentation with suitable replacement classifiers did not yield a candidate algorithm that was as accurate as the ridge regressor for ROCKET.

\begin{figure}[]
    	\centering
        \includegraphics[width=\linewidth,trim={2cm 9cm 0cm 4cm},clip]{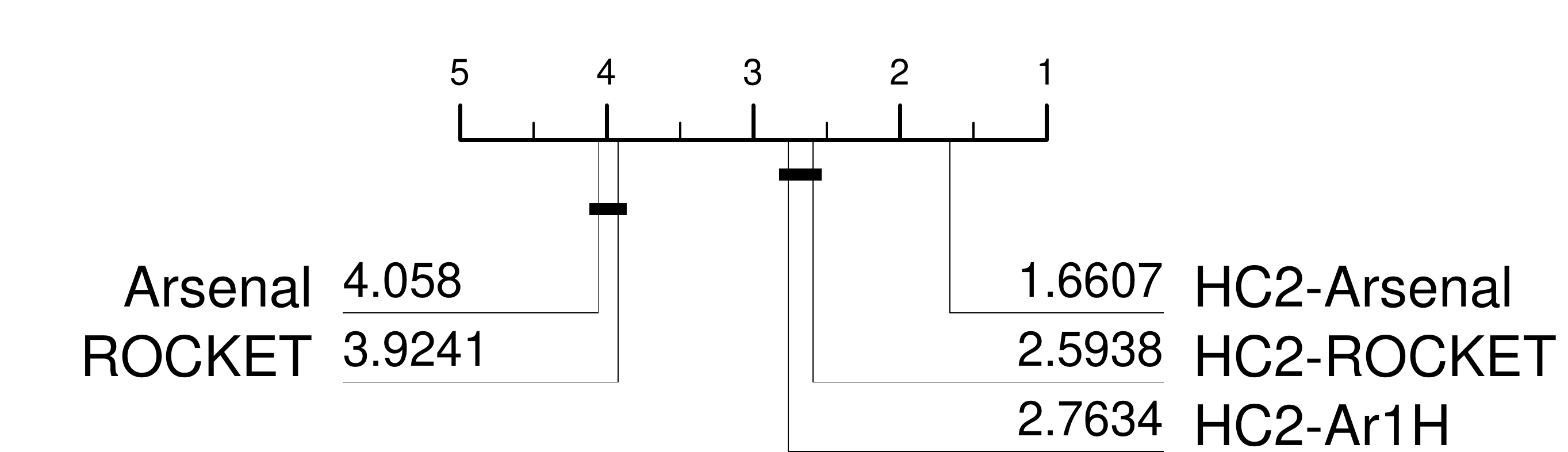}
        \caption{Critical difference diagram for both versions of ROCKET and versions of HIVE-COTE using them on 112 UCR datasets. HC2-Ar1H represents HIVE-COTE using the Arsenal classifier with probabilities generated in the same way as ROCKET.}
        \label{fig:arsenalCD}
    \end{figure}

    To solve this problem, the version of ROCKET we use in HIVE-COTE is an ensemble of smaller ROCKET classifiers. We refer to this fusillade of ROCKETs as the Arsenal. New cases are classified using a majority vote. Arsenal is slower to build than ROCKET, but its improved probabilities make it a better candidate for HC2. The build process for Arsenal is described in Algorithm~\ref{alg:Arsenal}. Figure~\ref{fig:arsenalCD} shows both versions of ROCKET and three versions of HC 2.0, two with a single version of both included in the ensemble and one containing a version of Arsenal where the probability of the predicted class is set to 1 rather than generated through the ensemble. Arsenal makes no improvement over default ROCKET in terms of accuracy, and Arsenal using the same method for generating probabilities as ROCKET makes no improvement in HC 2.0. However, the HC 2.0 version including an unaltered Arsenal is significantly better. Even with probabilities estimated through the votes of a small sized ensemble, a large difference is made over having none at all in HIVE-COTE.

    \begin{algorithm}[htb]
        \caption{Arsenal(A list of $n$ cases of length $m$ with $d$ dimensions, ${\bf T}=({\bf X,y})$)}
        \label{alg:Arsenal}
        \begin{algorithmic}[1]
            \REQUIRE the ensemble size, $r$ and the number of kernels per classifier, $k$ (default r=25 and $k=2000$)
            \STATE Let $\bf F$ be a list of ROCKET classifiers $({\bf f}_1,\ldots,{\bf f}_r)$
        	\FOR {$i \leftarrow 1$ to $r$}
        	    \STATE Let $\bf X'$ be a $n$ by $2k$ list of transformed cases
        	    \FOR {$j \leftarrow 1$ to $k$}
        	        \STATE $[l,w,b,a,p,O] \leftarrow randomKernelParameters(m,d)$
        	
        	        \FOR {$t \leftarrow 1$ to $n$}
        	            \STATE $max \leftarrow -\infty$
        	            \STATE $ppv \leftarrow 0$
        	
        		        \FOR {$g \leftarrow 1$ to $(m+(2p))-((l-1)a)$}
            		        \STATE $v \leftarrow 0$
        		            \FOR {$c \leftarrow 1$ to $|O|$}
        		                \STATE $v \leftarrow v + applyKernel($$\bf X$$_t,g,O_c,l,w,b,a)$
        		            \ENDFOR
        		            \IF{$v > max$}
                                \STATE $max \leftarrow v$
            		        \ENDIF
        		            \IF{$v > 0$}
            		            \STATE $ppv \leftarrow ppv + 1$
            		        \ENDIF
        		        \ENDFOR
        		
        		        \STATE $\bf X'$$_{t,2j-1} \leftarrow max$
        		        \STATE $\bf X'$$_{t,2j} \leftarrow ppv / ((m+(2p))-((l-1)a))$
        		    \ENDFOR
        	    \ENDFOR
        	
        	    \STATE $f_i.buildRidgeClassifierCV([$$\bf X'$$,y])$
        	\ENDFOR
        \end{algorithmic}
    \end{algorithm}

\subsection{Shapelet Transform Classifier (STC)}
Shapelets are phase independent subseries found in the training data. The STC approach to classification using shapelets is to construct a pipeline where the search for high quality shapelets is followed by a transformation where the new features represent distances to retained shapelets. A rotation forest~\citep{rodriguez06rotf} is constructed on the transformed features. The shapelet transform is highly configurable: it can use a range of sampling/search techniques in addition to alternative quality measures. We present the default settings and direct the interested reader to the \texttt{tsml} code. The original shapelet based algorithms performed an exhaustive search of all possible shapelets. This of course is very slow. However, subsequent work~\citep{bostrom17binary} identified that exhaustive search can actually lead to over fitting and is never necessary. Instead, we randomly search for shapelets for a given amount of time, which is now a parameter (defaults to one hour). Our version of STC is essentially the same as that used for HC1~\citep{bagnall20hivecote1}, so we direct the interested reader there for more details. The multivariate version searches dimensions independently and is the same version used in the MTSC bake off~\citep{ruiz21mtsc}.

\section{Experimental Structure}
\label{sec:methodology}

    We perform our univariate time series experiments on 112 of the 128 datasets from the UCR time series archive~\citep{dau19ucr}. We exclude datasets containing series of unequal length or missing values, as we do not want an algorithm's aptitude for these cases to alter results and most implementations are not set up to handle these kinds of data. We additionally remove the Fungi data, which only provides a single train case for each class label.
    For our multivariate experiments we use all 26 equal length datasets of the 30 total from the UEA multivariate time series archive~\citep{bagnall18mtsc}.
    For each dataset we present performance as an average over 30 resamples. Both archives provide a default split into train and test sets which we use for the first resample. The remaining 29 are randomly resampled from the original split in a stratified manner.
    We seed each classifier and data resample using the fold index to ensure out results are reproducible.

    All of our non-deep learning experiments were run using the Java \texttt{tsml} toolkit implementations.
    For deep learning approaches we use Python \texttt{sktime} companion package \texttt{sktime-dl}\footnote{https://github.com/sktime/sktime-dl}. The configuration for each algorithm is provided in Table~\ref{tab:classifiers}.

    \begin{table}[]
        \caption{Classifier configurations for our experiments where $m$ is the series length, $d$ is the number of dimensions and $rm$ is the lengths of DrCIF representations.}
        \centering
        \label{tab:classifiers}
        \begin{tabular}{l|l}
            \hline
            Algorithm         &  Configuration \\ \hline
            TDE               & 250 parameter sets sampled, 50 max ensemble size \\
            DrCIF             & 500 trees, 4+(sqrt($rm$)*sqrt($d$))/3 intervals per representation, \\
                              & 10 attributes per tree, $rm$/2 max interval length \\
            Arsenal           & 2,000 kernels per classifier, 25 ensemble size \\
            STC               & 1 hour Shapelet Transform train time contract, \\
                              & 200 Rotation Forest trees \\ \hline
            ROCKET            & 10,000 kernels \\
            HIVE-COTE 1.0     & $\alpha$ of 4 \\
            HIVE-COTE 2.0     & $\alpha$ of 4 \\
            TS-CHIEF          & 500 trees, 5 EE splitters, 100 RISE splitters, 100 BOSS splitters \\
            InceptionTime     & Epochs 1500, batch size 64, learning rate $1e-3$ and halved after \\                                      & no improvement for 50 epochs. \\
                              & Two residual blocks each with three Inception modules with \\
                              & kernel sizes per module [10, 20, 40] plus bottleneck filters for all \\
                              & conv layers of 32 \\ \hline
            CIF               & 500 trees, sqrt($m$)*sqrt($d$) intervals, 8 attributes per tree, \\
                              & $m$/2 max interval length \\
            DTW\_D            & Full warping window \\
            \hline
        \end{tabular}
    \end{table}

    Our experiments using algorithms from \texttt{tsml} were conducted on the UEA high performance computing (HPC) cluster. Each job consists of a single dataset, classifier, fold evaluation and runs on a single core. Due to limits on the cluster, a job has a maximum  run time of seven days. The maximum memory allowance provided by the cluster stands at 500GB.

    %, with both these factors limiting what is feasible to include in our experimentation. HIVE-COTE 2.0 surpasses these limits on some datasets, this is possible as we run each component of HIVE-COTE 2.0 individually and post-process it from results files.
    \texttt{sktime-dl} experiments were performed on desktops GPUs, one with a Titan XP and one with 4 Titan X Pascals. Each job is run on a single GPU with each GPU running only one job at a time. There is no time limit for running these jobs. However, they are limited by the GPU memory of 12GB per card.

    We evaluate classifier performance using accuracy, area under the receiver operator curve (AUC) and negative log-likelihood. This means we can assess classifiers based on predictive performance, ranking predictions and probability estimates. For problems with more than two classes, one-vs-many AUC is averaged over the class values, weighted by class value frequency.

    To compare the scores averaged over 30 resamples of two classifiers for multiple datasets, we use a pairwise Wilcoxon signed-rank test. For multiple classifiers over multiple datasets we use an adaptation of the critical difference diagram~\citep{demsar06comparisons}, replacing the post-hoc Nemenyi test with a comparison of all classifiers using pairwise Wilcoxon signed-rank tests, and cliques formed using the Holm correction recommended by~\citep{garcia08pairwise,benavoli16pairwise}.

\section{Results}
\label{sec:results}
We have conducted extensive experimentation and produced a high volume of results. There is marginal utility in having massive tables of results. Instead, we present summarised results and have put the raw results and all of the summary analysis  on the accompanying website\footnote{www.timeseriesclassification.com/HC2.php}. Details of how to reproduce the results, or some subset of results, are given on the website and the code walk through on the \texttt{tsml} github.

    \subsection{Performance on the UCR Archive of 112 Univariate TSC Problems}

\begin{figure}[htb]
    	\centering
        \includegraphics[width=\linewidth,trim={1cm 4cm 0cm 4cm},clip]{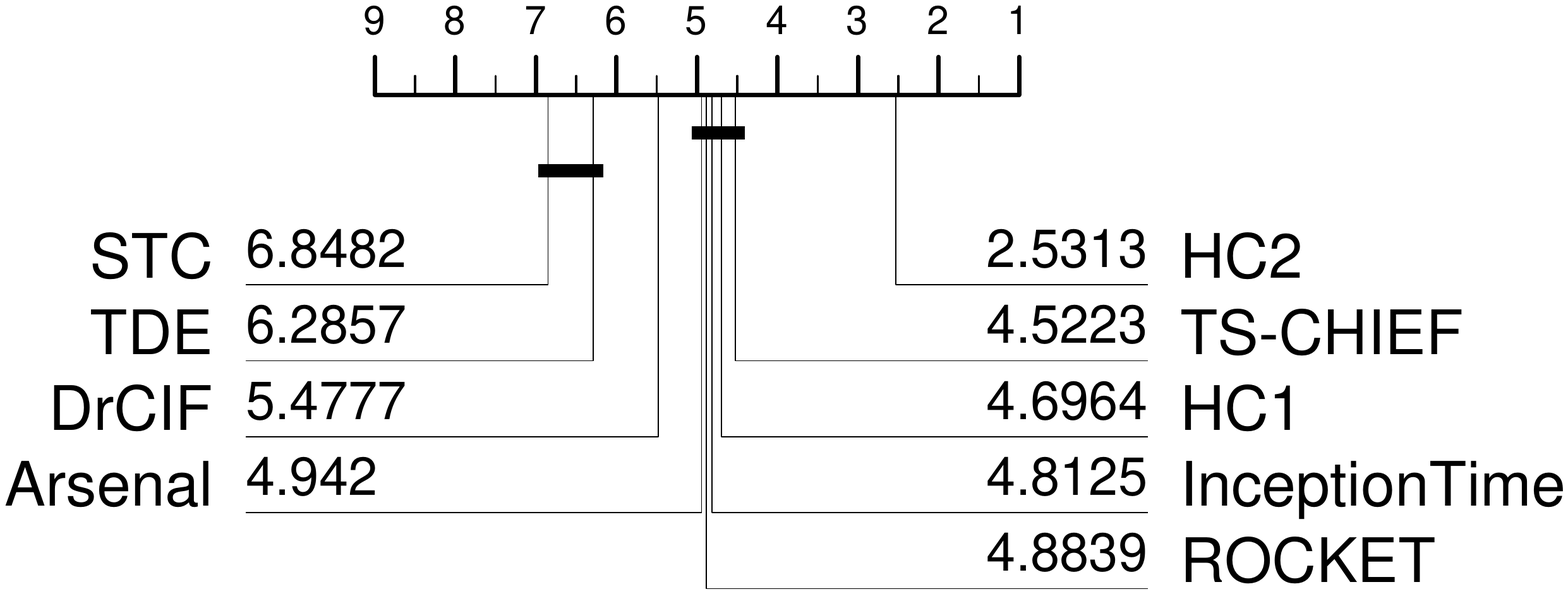}
        \caption{Test accuracy critical difference diagram for nine classifiers, averaged over 30 resamples for each of the 112 UCR problems. }
        \label{fig:sota-acc}
    \end{figure}

\begin{figure}[htb]
    	\centering
        \includegraphics[width=\linewidth,trim={1cm 4cm 0cm 4cm},clip]{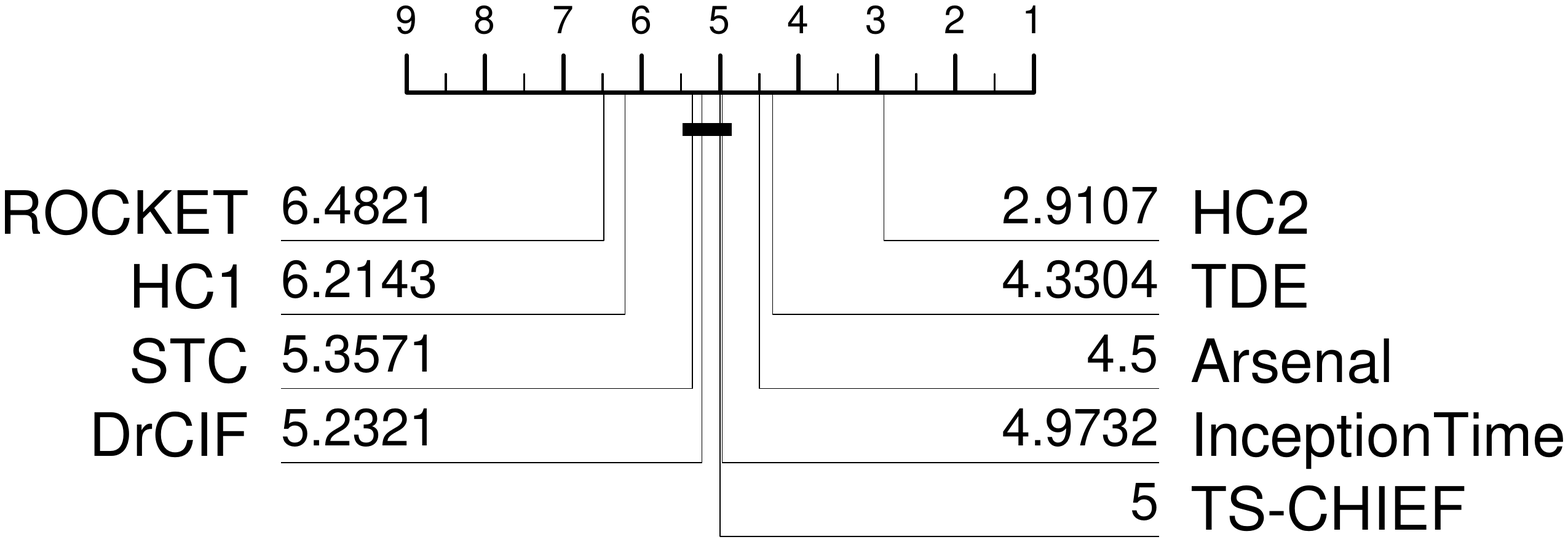}
        \caption{Test negative log likelihood critical difference diagram for  nine classifiers, averaged over 30 resamples for each of the 112 UCR problems. }
        \label{fig:sota-nll}
    \end{figure}

\begin{figure}[htb]
    	\centering
        \includegraphics[width=\linewidth,trim={1cm 4cm 0cm 4cm},clip]{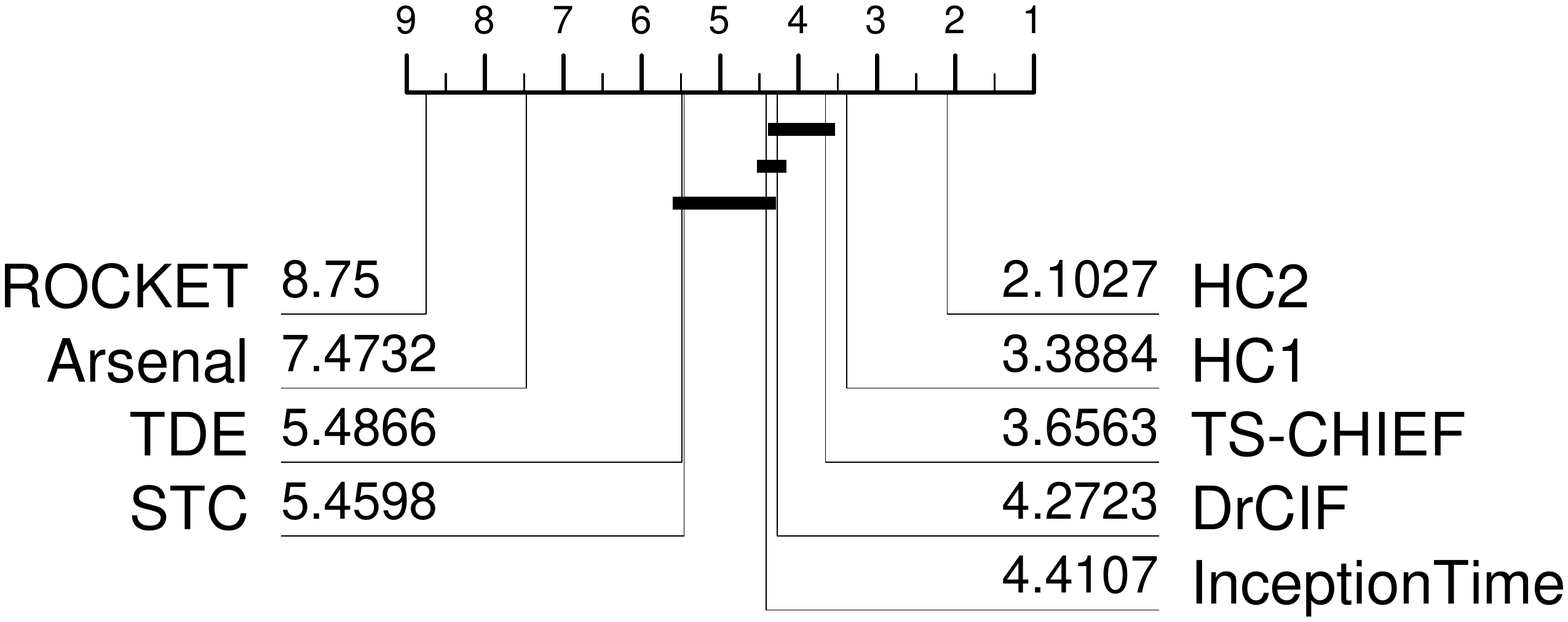}
        \caption{The area under the receiver operator curve critical difference diagram for  nine classifiers, averaged over 30 resamples for each of the 112 UCR problems.}
        \label{fig:sota-auroc}
    \end{figure}

Our core result is that HC2 is significantly better than the current state of the art on the 112 UCR equal length datasets. Figure~\ref{fig:acc-cd} in the Introduction shows the accuracy performance of HC2 vs the four baseline approaches.  Figures~\ref{fig:sota-acc},~\ref{fig:sota-nll} and~\ref{fig:sota-auroc} shows the critical difference diagrams for accuracy, NLL and AUROC for HC2, its four components and the four state-of-the-art algorithms. They demonstrate that HC2 is significantly better than its components and the four benchmarks using all three metrics. For accuracy (Figure~\ref{fig:sota-acc}), STC and TDE form the lowest clique. DrCIF is between these and the current state of the art, which contains Arsenal. ROCKET is the worst for probability estimates  (Figure~\ref{fig:sota-nll}), since it only produces 0/1 estimates. The better probability estimates of Arsenal justify our design decision, since HC weights distributions not predictions. TDE is surprisingly good at probabilistic prediction. The poor probability estimates of HC1 highlights one source of improvement in HC2. HC1 does much better at AUROC (Figure~\ref{fig:sota-auroc}), whereas Arsenal does much worse, indicating further calibration may benefit the Arsenal.


\begin{figure}[!ht]
	\centering
\begin{tabular}{cc}
       \includegraphics[width =6.5cm, trim={3cm 0cm 2cm 0cm},clip]{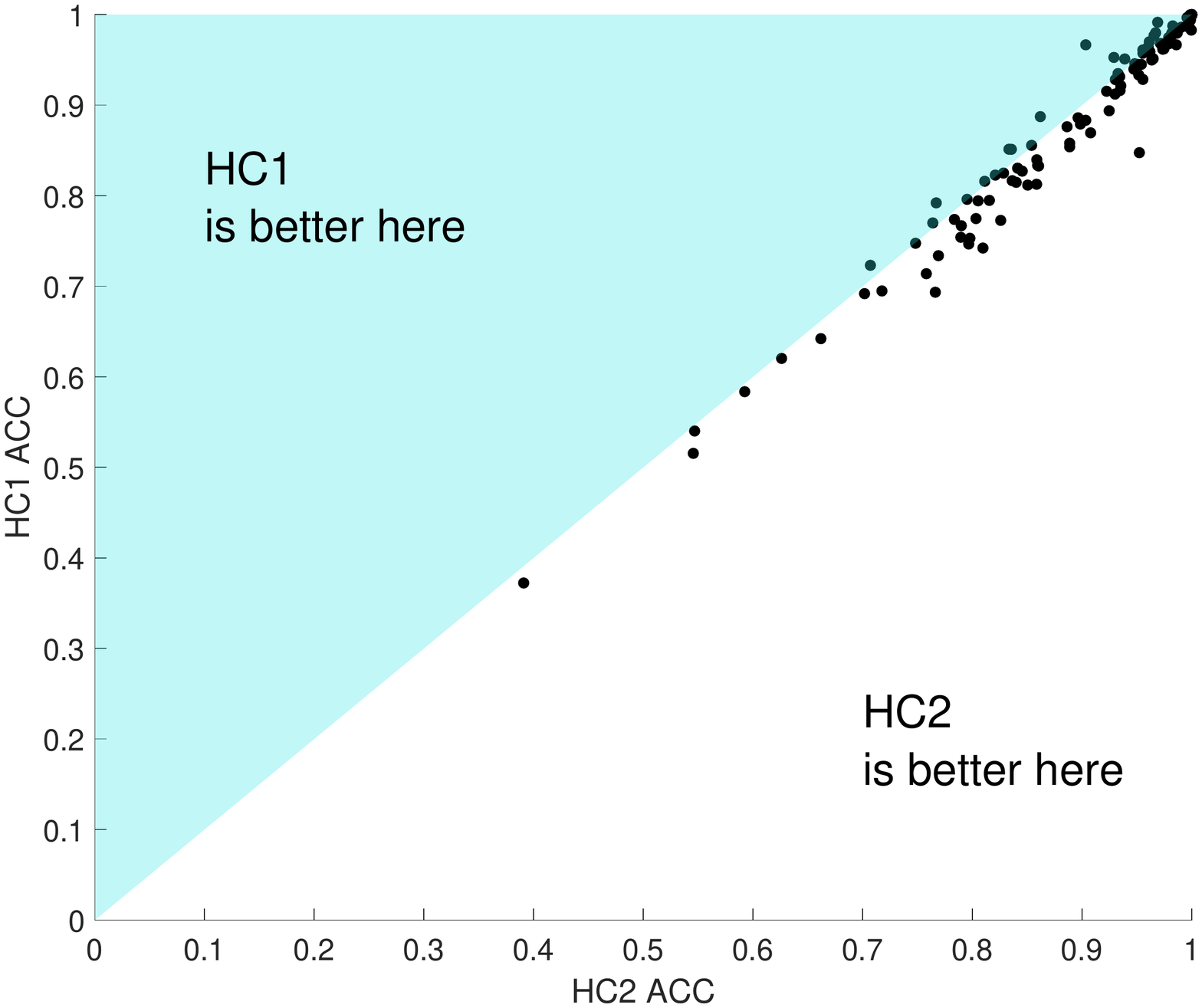}              	
&
       \includegraphics[width =6.5cm, trim={3cm 0cm 2cm 0cm},clip]{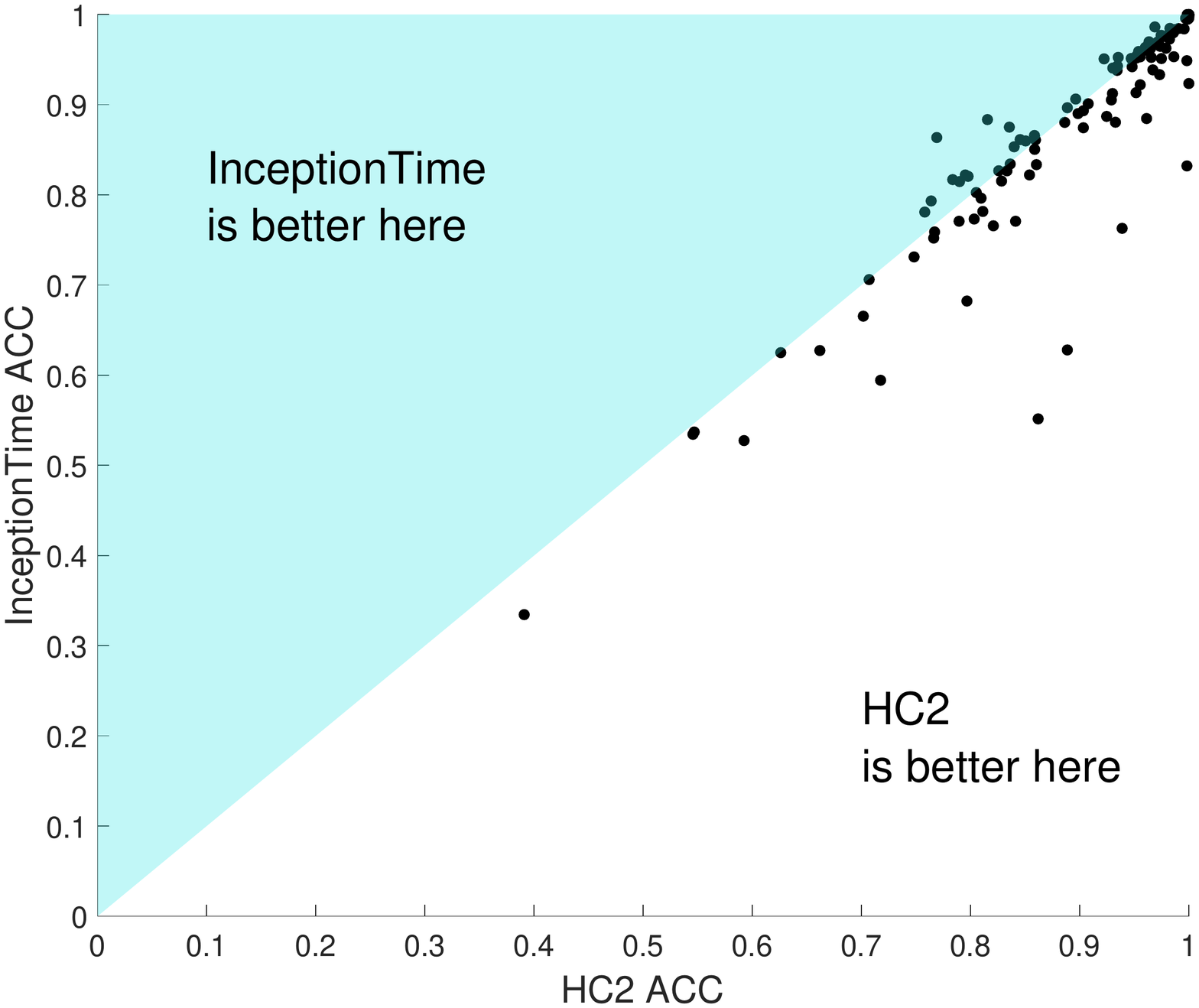}  \\

       \includegraphics[width =6.5cm, trim={3cm 0cm 2cm 0cm},clip]{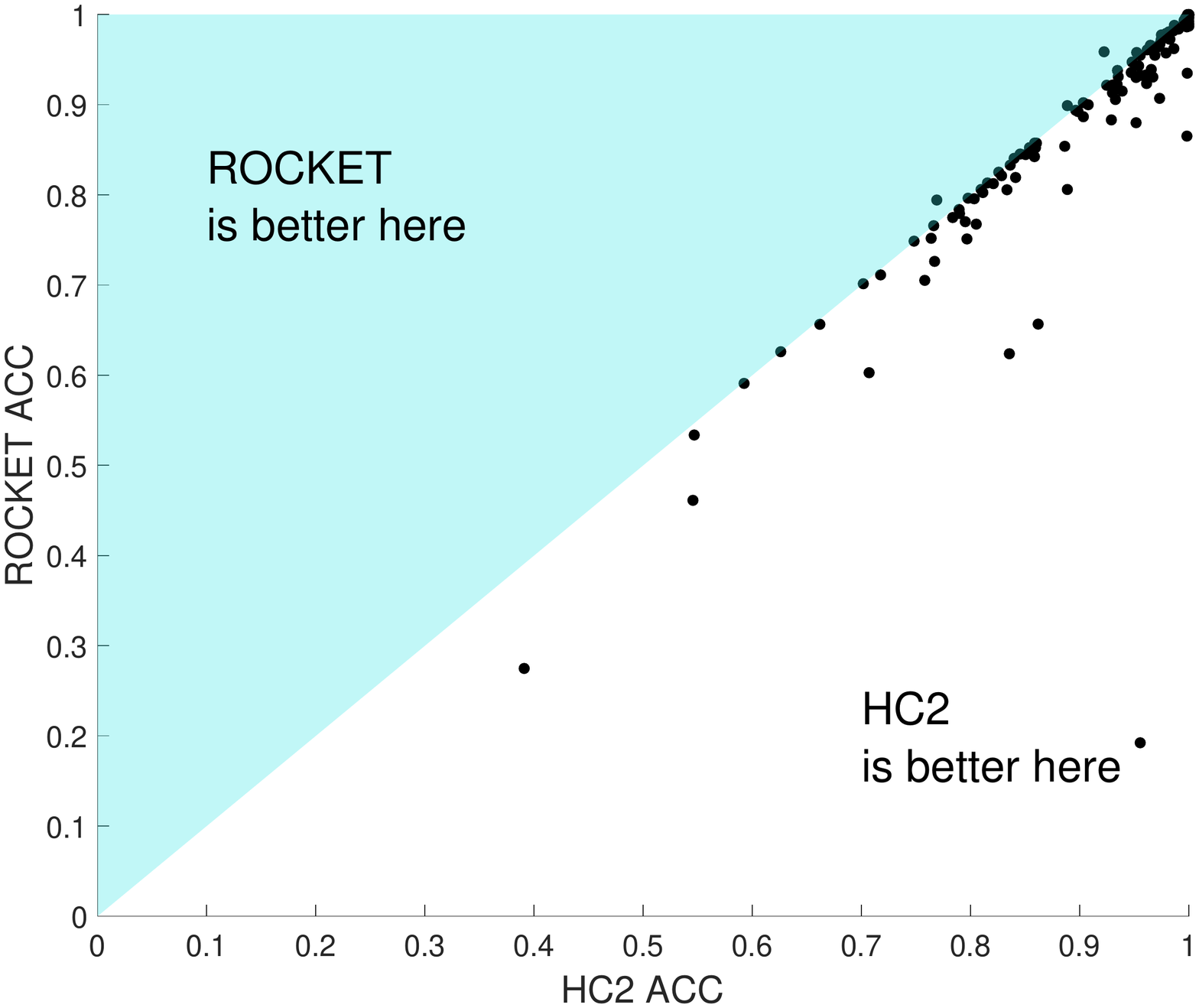}              	
&
       \includegraphics[width =6.5cm, trim={3cm 0cm 2cm 0cm},clip]{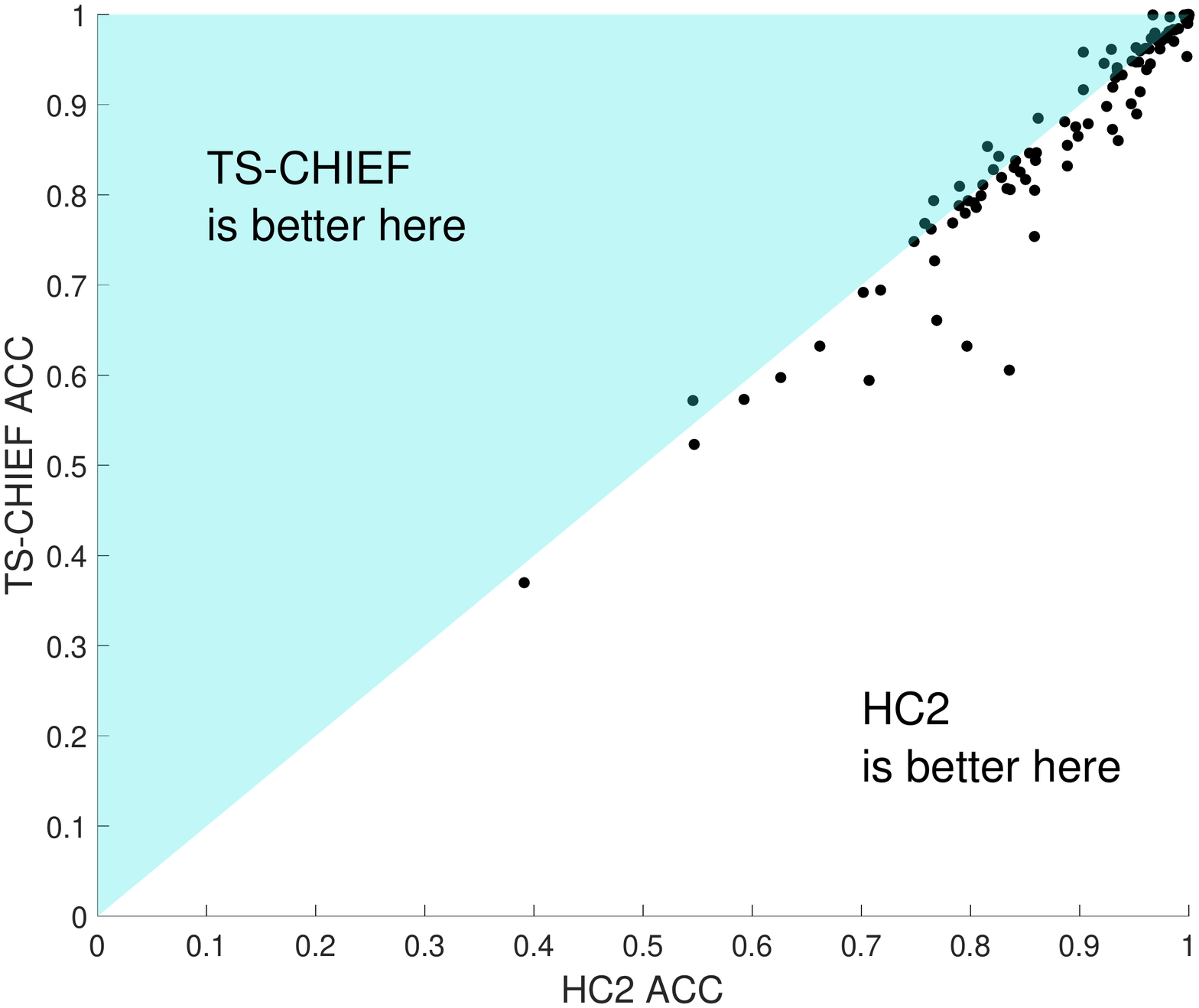}  \\

       \end{tabular}
       \caption{Scatter plots of HIVE-COTE 2.0 against each of the baseline classifiers}
       \label{fig:sota-scatter}
\end{figure}

Figure~\ref{fig:sota-scatter} shows the accuracy scatter plots of HC2 against each of baseline classifiers and Table~\ref{tab:sota-summary} summarises the differences in test accuracy between HC2 and the four baselines.

%Critical difference diagrams CAWPE-S with its base classifiers (left), and CAWPE-A with its base classifiers (right). Ranks formed on test set accuracy averaged over 30 resamples.

\begin{table}[htb]
    \centering
    \caption{Summary of the differences between HC2 and the benchmarks. A negative value means the HC2 is better. }
    \label{tab:sota-summary}
    \begin{tabular}{c|c|c|c|c|c|c|c}
       Classifier   & Mean      & Median    & Max        & Min     &  StDev     & HC2 Wins & HC2 Loses  \\ \hline
TS-CHIEF	        & -1.36\%   &	-0.41\% &	-22.99\% &	5.50\% & 3.64\%     & 77	  & 29 \\
InceptionTime	    & -1.69\%   &	-0.37\% &	-31.04\% &	9.46\% & 5.13\%	        & 74	  & 32 \\
HC1	                & -1.06\%    &   -0.69\% &	-10.47\% &	6.33\% & 2.10\%	        & 82	  & 25 \\
ROCKET	            & -2.49\%   &	-0.72\% &	-76.31\% &	3.64\% & 7.92\%	        & 97	  & 11   \\ \hline
\end{tabular}
\end{table}
We observe that there is lower variance between HC1 and HC2, but that HC2 consistently outperforms HC1 with an average accuracy of more than 1\%. The variation in difference to HC2 is greater with the other three classifiers, in particular ROCKET. The median difference is lower than the mean in all cases. This suggests skew, which supports the core hypothesis that the heterogenous ensemble can compensate for the shortcomings of its components. It also suggests that HC2 has a higher representational power, in that it can find a more diverse set of features.

Accuracy is not the only consideration. Table~\ref{tab:runtime} summarises the run time and memory requirements for the classifiers compared in Figure~\ref{fig:sota-scatter}. There are a few caveats to these results. Firstly, all of the results except InceptionTime are run in a single thread on a CPU. Thus InceptionTime time experiments are not really directly comparable, since it runs on a GPU. ROCKET and HC2 are forced to run in a single thread, despite being threadable. The times for the HC2 components are without the time to estimate performance, but these are included in the HC2 times. Memory is the maximum memory used throughout the run, as obtained from the Java garbage collector, and should be considered approximate. We are not set up to measure the maximum memory used with InceptionTime in practice, but we know it did not exceed 12GB, because that was the memory available on the GPU. Since the run times are sequential, we also use the sequential memory for HC1 and HC2. These would be higher if the classifier were threaded, but of course the run time would be much lower.
\begin{table}[htb]
    \centering
    \caption{Run time and memory requirements to train single resample of 112 UCR problems. The median of 30 runs is taken for each dataset. }
    \label{tab:runtime}
    \begin{tabular}{c|c|c|c}
 Algorithm       &  Total (hrs) & Average (mins) & Max Mem (MB) \\ \hline
ROCKET	        & 2.85    &     1.53     & 4349               \\
Arsenal	        & 27.91   &    14.95    & 1683 \\
DrCIF	        & 45.40   &   24.32     & 920 \\
TDE	            & 75.41   &     40.40   & 6565\\
InceptionTime	& 86.58   &   46.38     &  - \\
STC	            & 115.88  &    62.08    & 4219\\
HC2	            & 340.21  &    182.26   & 6677 \\
HC1	            & 427.18  &   228.84    & 4876 \\
TS-CHIEF	    & 1016.87 &       544.75 & 26052 \\
    \end{tabular}
\end{table}

\begin{figure}[htb]
    	\centering
        \includegraphics[width=\linewidth,trim={2cm 4cm 2cm 4cm},clip]{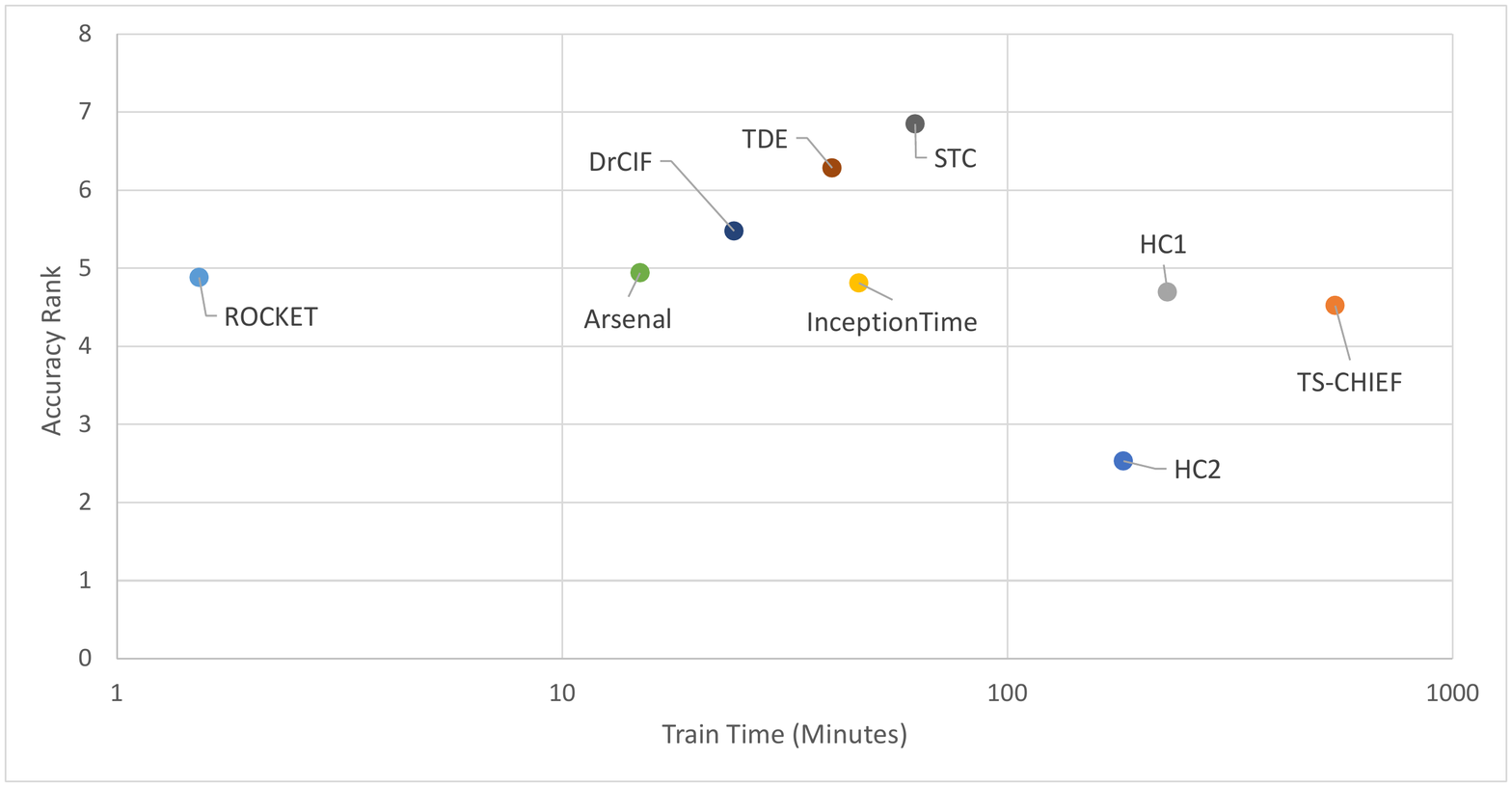}
        \caption{A comparison of classifiers in terms of accuracy rank and train time. The time and accuracy are averaged over 112 UCR problems. The train time is on a log scale. }
        \label{fig:rankvtime}
    \end{figure}

With this in mind, we can make the following observations. ROCKET lives up to its name and can build models on all 112 data in under 3 hours, even when not threaded. If speed is your main criteria, ROCKET is a good starting point in any analysis. STC is the slowest component, but this is caused by the configuration rather than an inherent problem: STC defaults to a one hour shapelet search or a full evaluation of the shapelet search if this will take less than an hour. For the very small problems, it takes a lot longer than the other algorithms (although still less than an hour). HC2 is faster than HC1, primarily because of improvements to STC and the change in classifiers. TS-CHIEF is the slowest algorithm by far, and seems to scale less well than the others. On the slowest five problems (HandOutlines, NonInvasiveFetalECGThorax1 and 2, SemgHandMovementCh2 and EthanolLevel), it takes ten times longer than HC2, but the difference is minimal on smaller problems.  All of the classifiers are within reasonable bounds for memory. TS-CHIEF has the highest memory requirement, with a max requirement of 26GB on HandOutlines. As with run time, it seems to scale worse than the others. HC2 requires more memory than HC1, but it is is not unreasonable. ROCKET has a worse max memory case (ElectricDevices) than Arsenal. Overall, ROCKET tends to use less memory than Arsenal but appears to scale worse for larger datasets with many cases. Arsenal uses a smaller amount of kernels for each individual classifier, meaning that each transformed set of data is smaller in size and discarded before the next is built. ROCKET on the other hand must transform using its larger amount of kernels at a single point. Figure~\ref{fig:rankvtime} summarises the accuracy and run time results by plotting the log of the train time against the rank.

    \subsection{Performance on the UEA Archive of 26 Multivariate TSC Problems}
\begin{figure}[htb]
    	\centering
        \includegraphics[width=\linewidth,trim={1cm 8cm 0cm 4cm},clip]{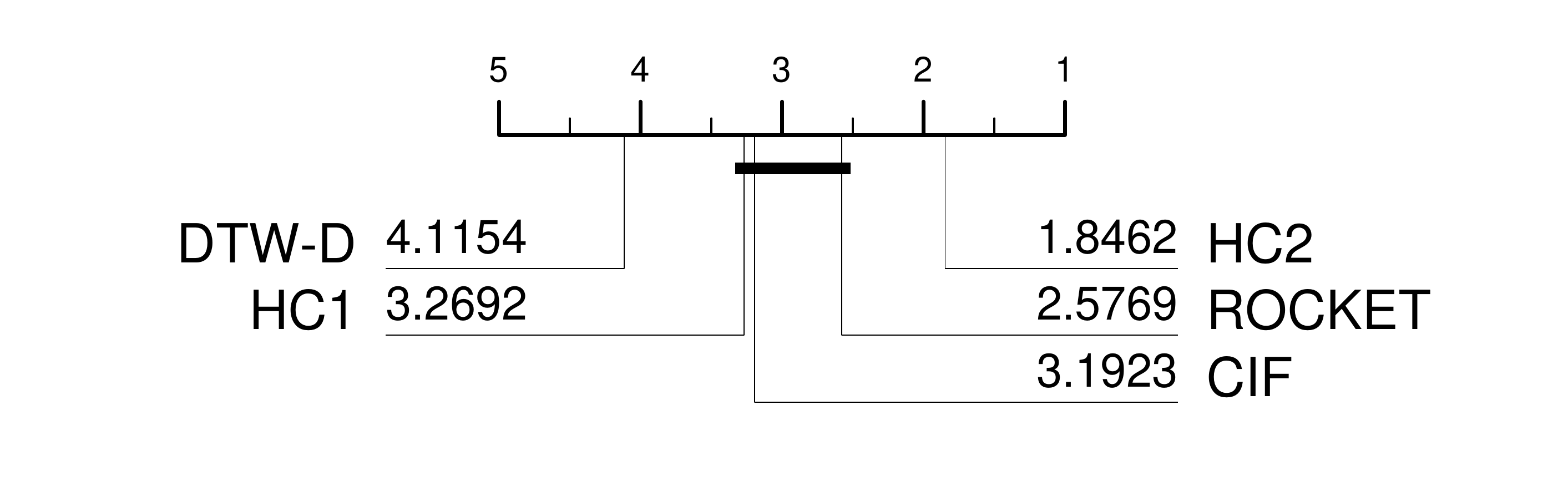}
        \caption{Test accuracy critical difference diagram for five classifiers, averaged over 30 resamples for each of the 26 UEA MTSC problems. }
        \label{fig:mstc-acc}
    \end{figure}

\begin{figure}[htb]
    	\centering
        \includegraphics[width=\linewidth,trim={1cm 8cm 0cm 4cm},clip]{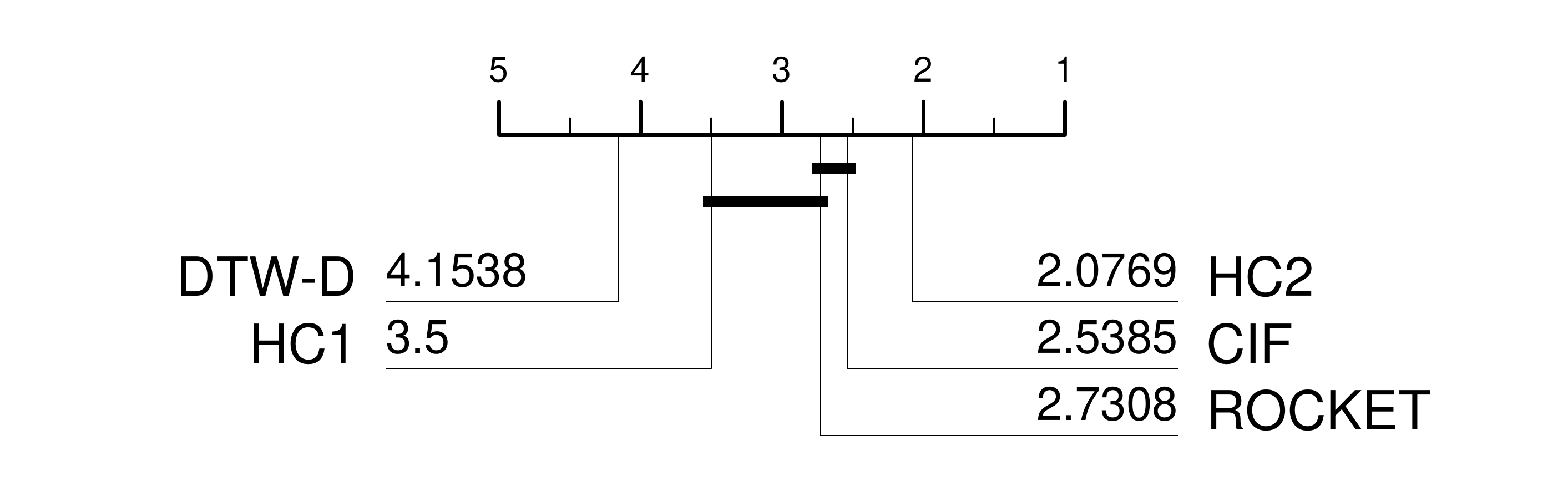}
        \caption{Test negative log likelihood critical difference diagram for five classifiers, averaged over 30 resamples for each of the 26 UEA MTSC problems.}
        \label{fig:mstc-nll}
    \end{figure}

\begin{figure}[htb]
    	\centering
        \includegraphics[width=\linewidth,trim={1cm 8cm 0cm 4cm},clip]{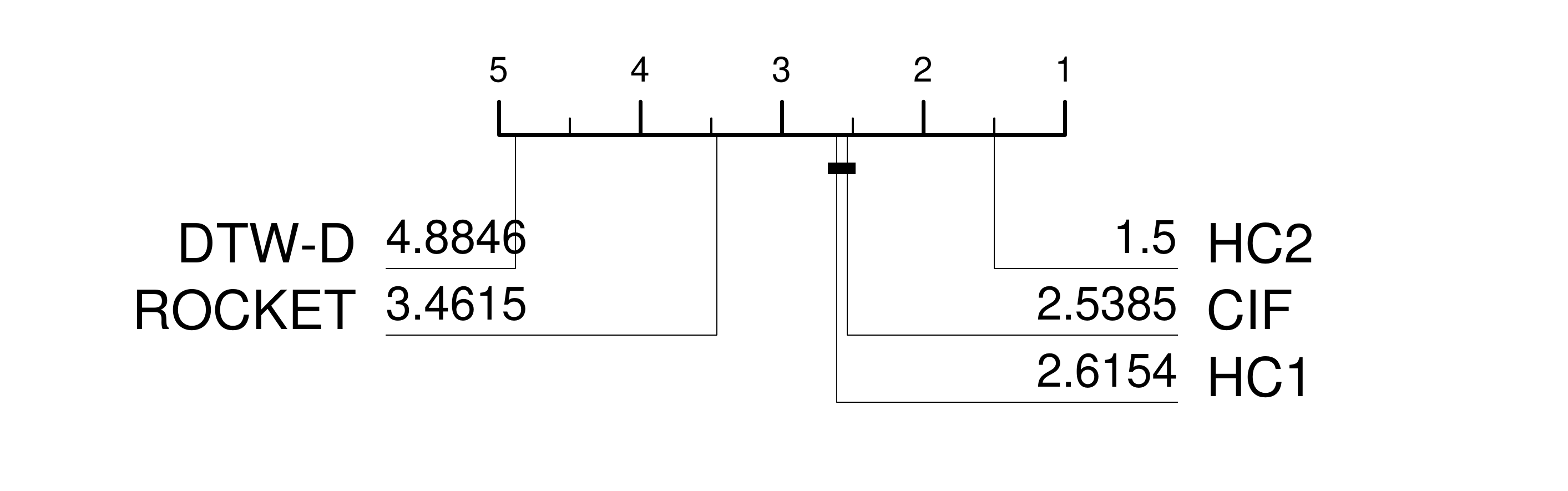}
        \caption{The area under the receiver operator curve critical difference diagram for five classifiers, averaged over 30 resamples for each of the 26 UEA MTSC problems.}
        \label{fig:mstc-auroc}
    \end{figure}

A bake off of TSC algorithms~\citep{ruiz21mtsc} using the MTSC UEA archive found three algorithms (that could complete all 26 data sets) were significantly more accurate than DTW-D. These were ROCKET, HC1 and CIF. We have repeated these experiments with HC2. Figures~\ref{fig:mstc-acc},~\ref{fig:mstc-nll} and~\ref{fig:mstc-auroc} show that HC2 is significantly better than DTW-D, ROCKET, HC1 and CIF on the 26 datasets for accuracy, NLL and AUROC.

\begin{figure}[!htb]
	\centering
\begin{tabular}{cc}
       \includegraphics[width =6.5cm, trim={3cm 0cm 2cm 0cm},clip]{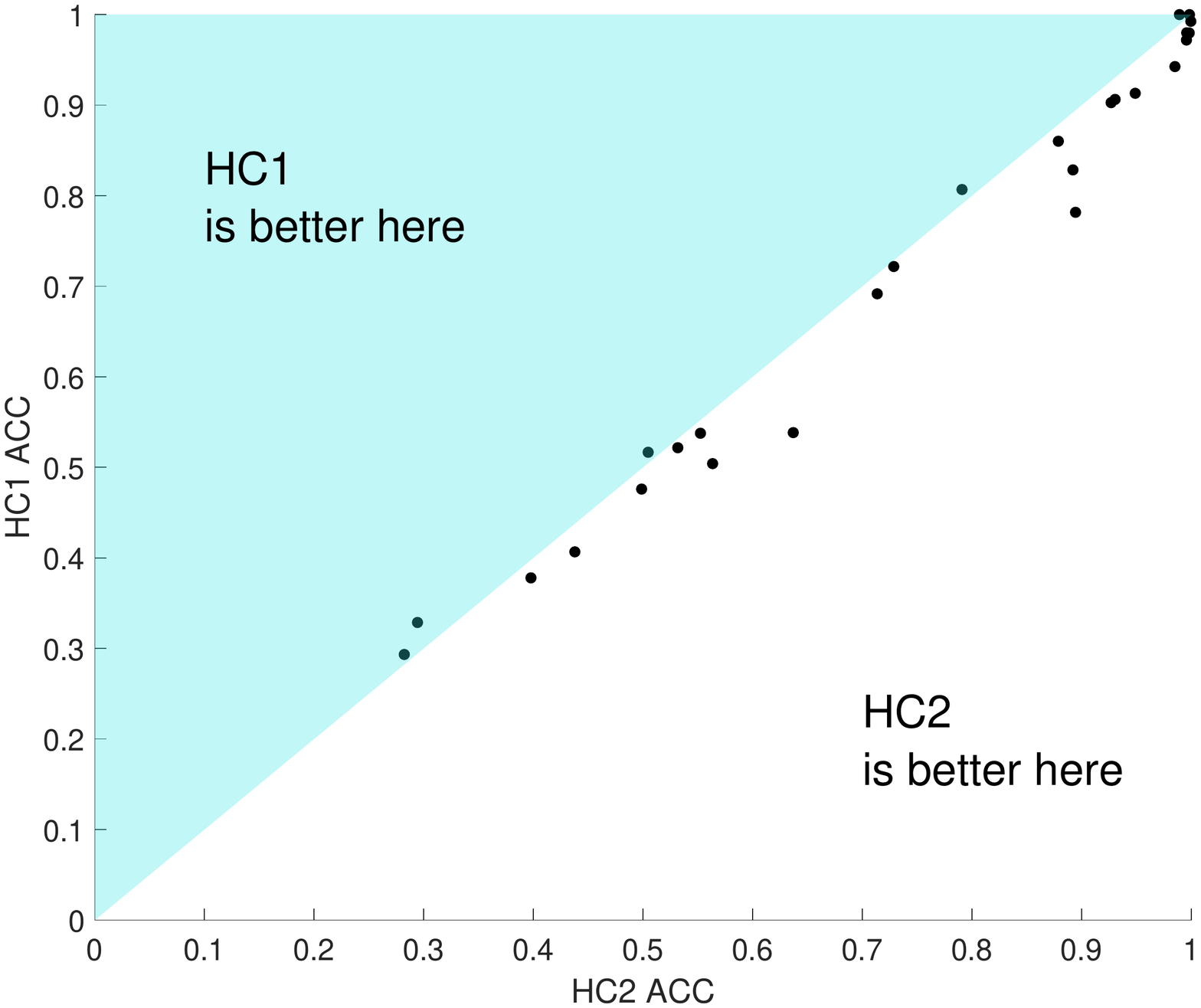}              	
&

       \includegraphics[width =6.5cm, trim={3cm 0cm 2cm 0cm},clip]{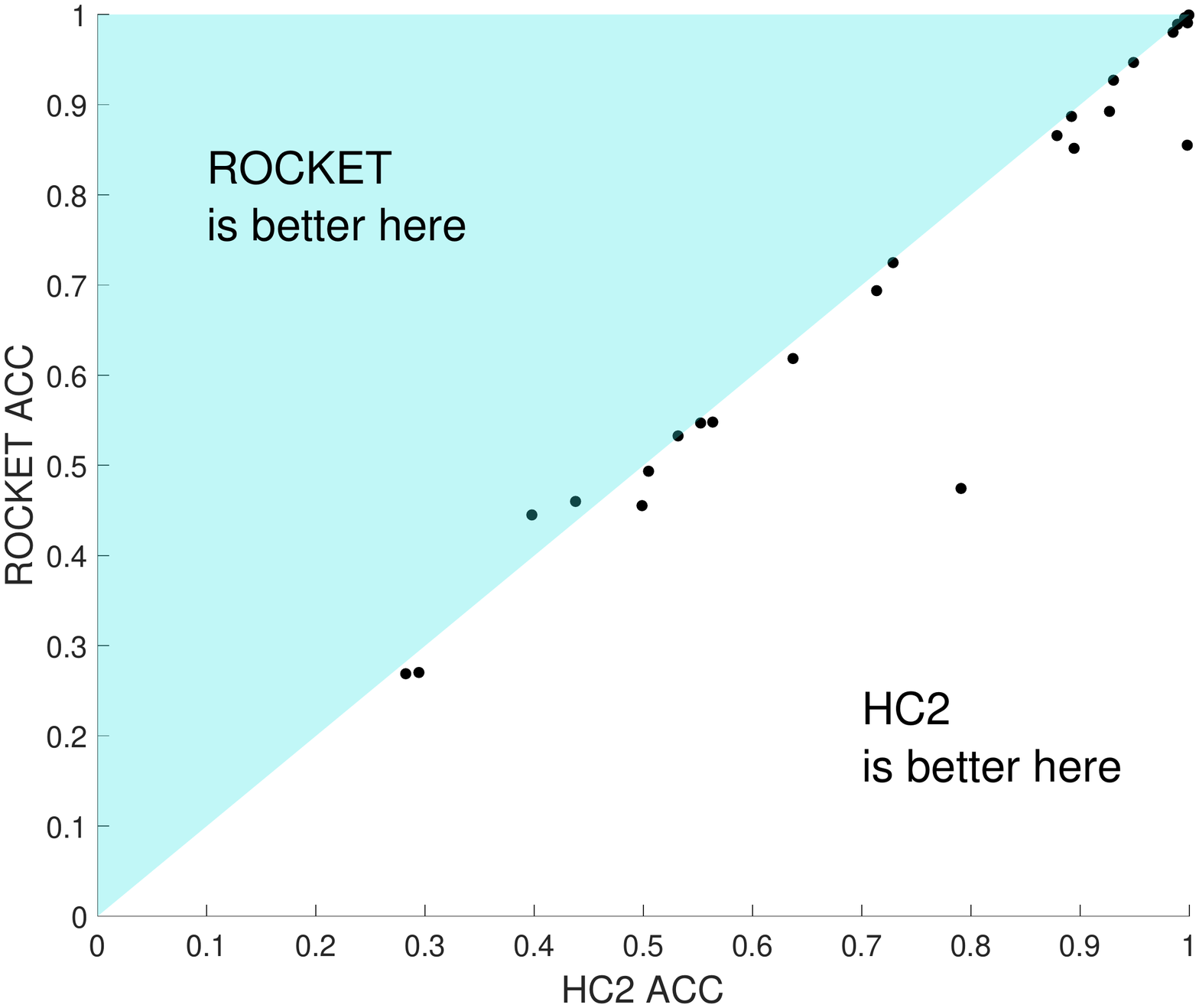}  \\            	

       \includegraphics[width =6.5cm, trim={3cm 0cm 2cm 0cm},clip]{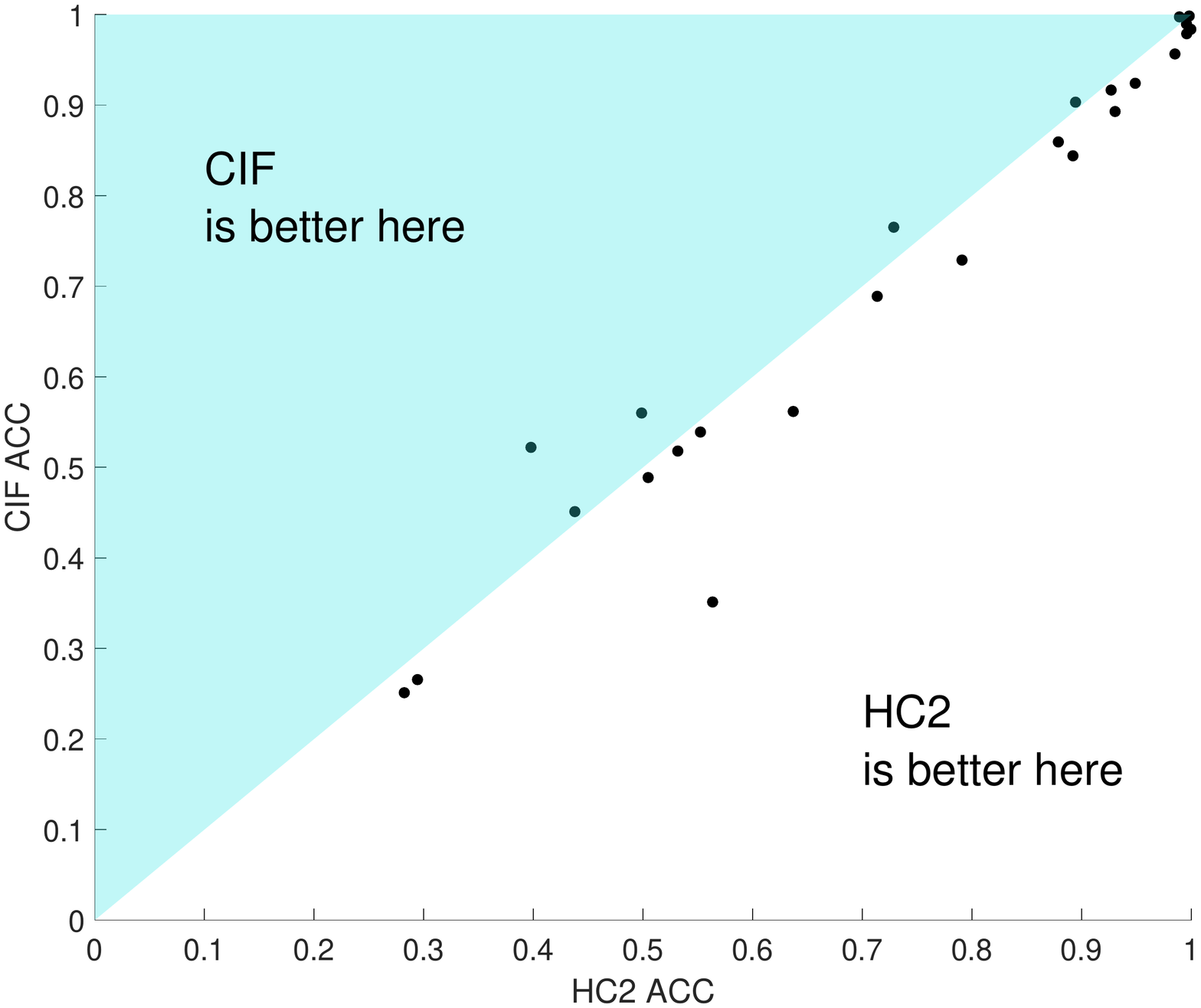}
&
     \includegraphics[width =6.5cm, trim={3cm 0cm 2cm 0cm},clip]{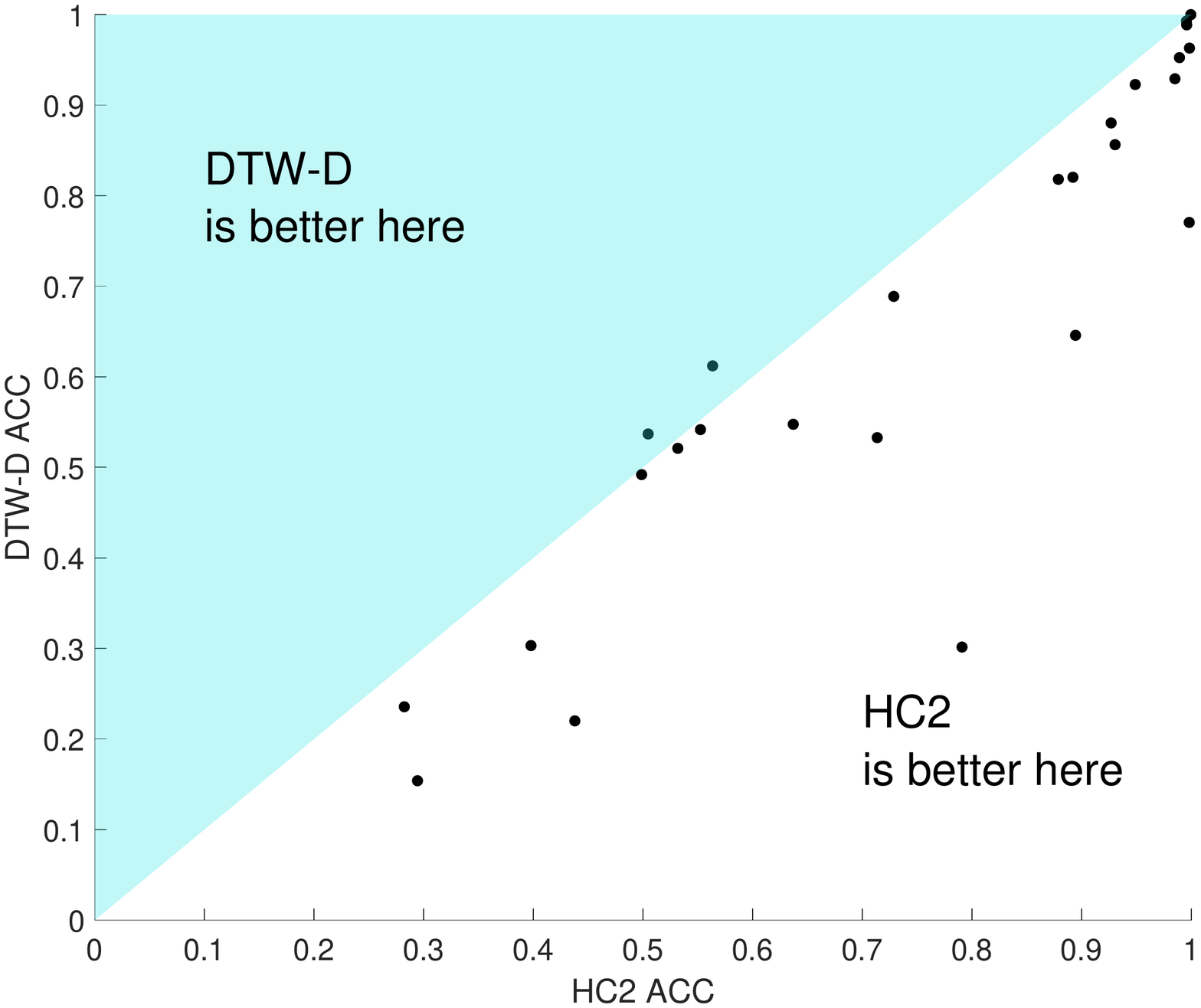}
       \end{tabular}
       \caption{Scatter plots of HIVE-COTE 2.0 against each of the baseline classifiers}
       \label{fig:mstc-scatter}
\end{figure}

Figure~\ref{fig:mstc-scatter} shows the accuracy scatter plots and Table~\ref{tab:mtsc-summary} summarises the differences  of HC2 against the benchmarks. We think these results strongly support the assertion that HC2 represents a new state of the art for multivariate time series classification.

\begin{table}[htb]
    \centering
    \caption{Summary of the differences between HC2 and the benchmarks. A negative value means the HC2 is better. }
    \label{tab:mtsc-summary}
    \begin{tabular}{c|c|c|c|c|c|c|c}
       Classifier   & Mean      & Median    & Max        & Min     &  StDev     & HC Wins & HC Loses  \\ \hline
ROCKET	    & -2.52\%  &	-0.64\% & -31.65\% &	4.73\% & 6.75\%   & 19	  & 5 \\
CIF	        &-1.71\%   & -1.58\% & -21.21\% & 12.43\% & 5.56\%   & 19	  & 7 \\
HC1         &-2.25\%   & -1.92\% & -11.27\% & 3.44\%  & 3.30\%   & 20	  & 6 \\
DTW-D	    & -8.22\%  &	-4.66\% & -48.94\% & 4.87\%  & 11.39\%  & 23      & 3   \\ \hline
\hline
\end{tabular}
\end{table}

% \subsection{Performance on new TSC Problems}
% We excluded unequal length problems from our study of the UCR data.

\section{Inside HC2: An Ablative Study}
 \label{sec:ablative}
    We address the question of why HC2 works so well, and evaluate design decisions made in the change from HC1 to HC2.
    HC1 uses cross validation to estimate the test accuracy from the train data for each component. HC2 modules are all ensembles, and so it was natural to attempt to use bagging and the out of bag accuracy estimate to speed up HIVE-COTE training. However, whilst this produces good estimates of the test accuracy, the models were less accurate on unseen data for every module. Hence, we made the decision to fit a separate bagging model for the estimation stage for those that need it, thus providing an order of magnitude speed up compared to cross validation. DrCIF and Arsenal both create separate models with bagging to generate their estimates. STC builds a new Rotation Forest model with bagging for its estimate, but uses the same transformed shapelet data for both. TDE naturally takes a 70\% subsample when creating its ensemble, as such a new model is not required to generate its out-of-bag error. However, we were concerned that these estimates may be biased and/or not consistent. Table~\ref{tab:estimates} summarises the distributions of the differences between estimated and observed test accuracy for HC2 and its components.

    \begin{table}[htb]
        \caption{Summary of the difference between estimated and observed test accuracy for HC2 and its components. A positive figure means that the classifier is overestimating accuracy from the train data.}
        \centering
        \begin{tabular}{c|c|c|c|c|c}
Classifier   &  Mean    & Median  &  Min  &	Max	& MSE\\ \hline
DrCIF	    & -2.15\% &	-0.93\% &	-46.78\% &	9.64\% &	0.40\% \\
Arsenal	    & -1.17\% &	-0.40\% &	-23.13\% &	9.23\% &	0.15\% \\
STC	        & 1.24\% &	0.78\% &	-55.54\% &	26.88\% &	0.86\% \\
TDE	        & -1.14\% &	-0.77\% &	-18.89\% &	10.02\% &	0.14\% \\
HC2	        & 0.47\% &	0.11\% &	-19.81\% &	19.17\% &	0.13\%
        \end{tabular}
        \label{tab:estimates}
    \end{table}

\begin{figure}[htb]
    	\centering
        \includegraphics[width=\linewidth,trim={1cm 9cm 2cm 11cm},clip]{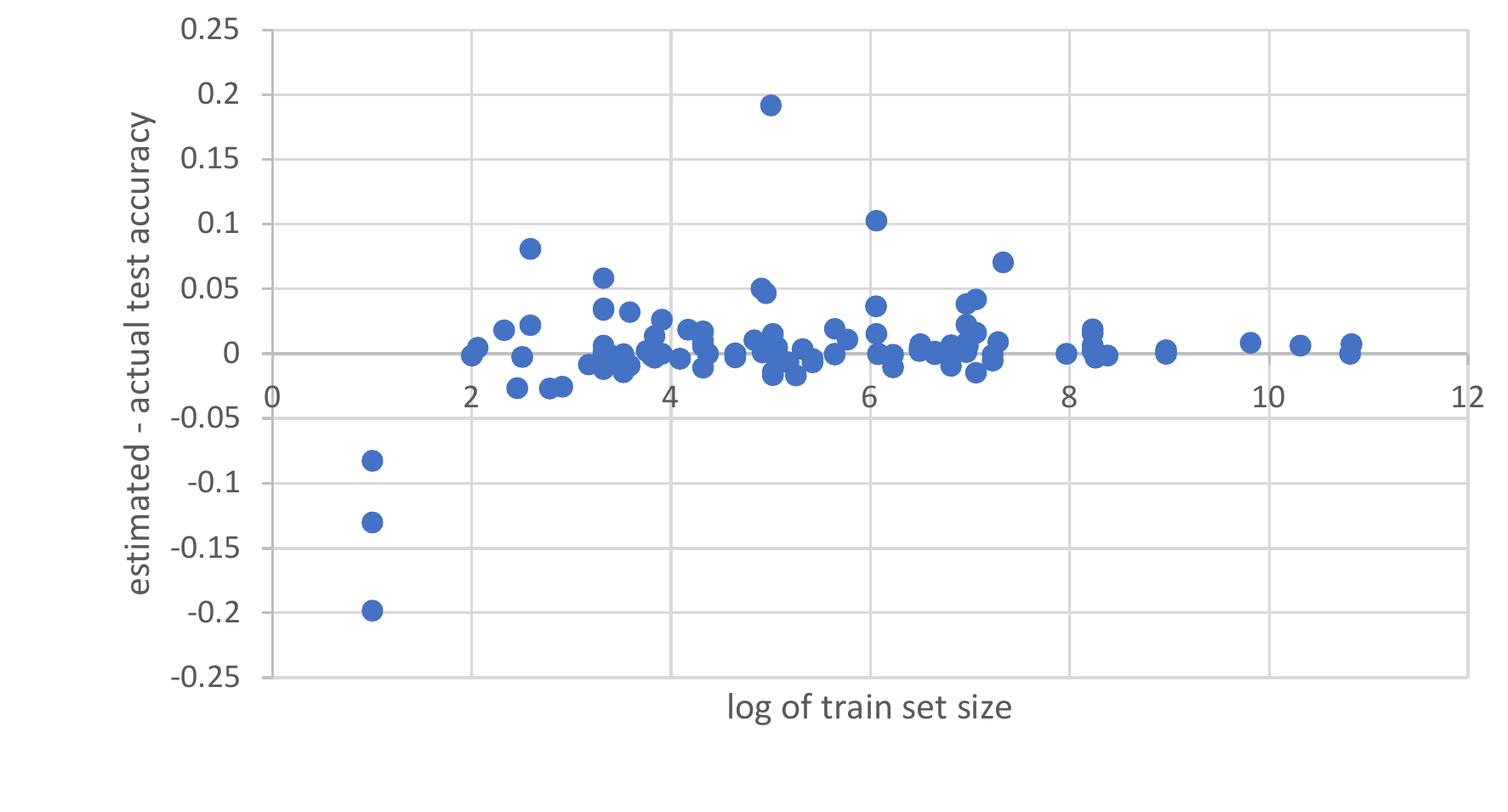}
        \caption{Difference in estimated and observed test set accuracy against the log of the train set size for 112 UCR datasets. }
        \label{fig:acc-est}
    \end{figure}

    Whilst there is small bias for each component, HC2 ensemble method compensates for this and has the lowest average deviation (and MSE deviation) between estimated and observed test accuracy. This is due to the averaging ensemble effect, and the biasing effect of reusing estimates from the components: a full nested cross validation estimate would be computationally demanding and is not necessary. STC is the only component that is over optimistic. This is to be expected. STC performs a random search on the whole train data then bags rotation forest. This introduces bias, and is a possible area for future improvement. The min and the max show that there  are some very large differences between estimate and observed. These primarily arise in problems where there are very few cases per class, such as PigAirwayPressure, PigCVP and PigArtPressure, which each have only two cases per class. Every classifier underestimates the test accuracy by over 10\% on these problems. Figure~\ref{fig:acc-est} shows the difference in the test accuracy estimate and actual plotted against the log of the train set size for HC2. The picture is not conclusive, but it could be argued that the variance of the difference is decreasing, which is encouraging evidence for the consistency of the HC estimate.

\begin{figure}[!htb]
    	\centering
        \includegraphics[width=0.85\linewidth,trim={2.5cm 6.1cm 2cm 7.5cm},clip]{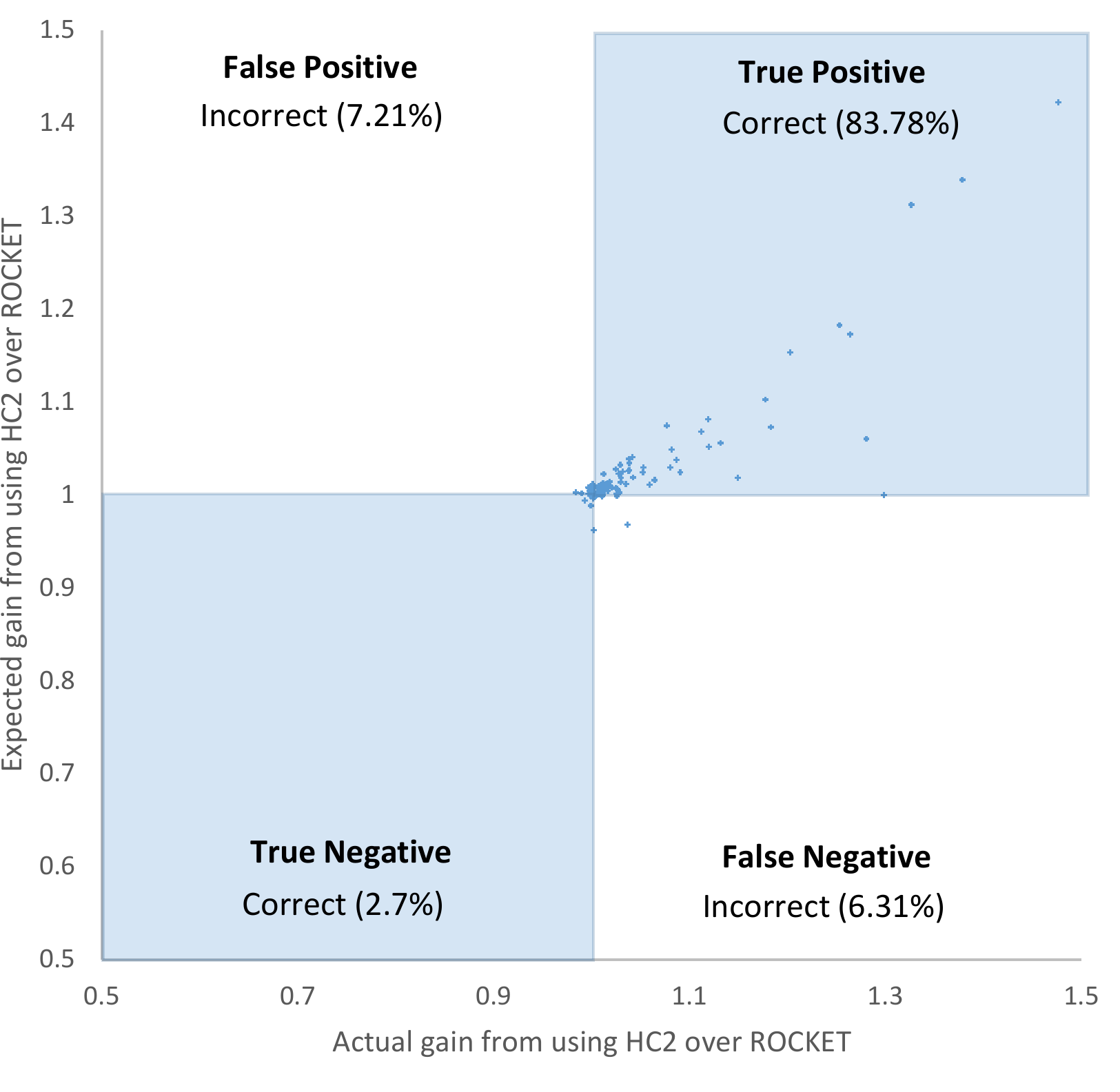}
        \caption{Texas sharpshooter plot for HC2 vs ROCKET. Each point represents a single dataset. The x-axis is the ratio of HC2 and ROCKET actual test accuracy and the y-axis is the ratio or predicted test accuracy.   }
        \label{fig:texas}
    \end{figure}

Another benefit of accurate estimates from the train data is that they can be used to compare classifiers with a Texas Sharpshooter plot~\citep{batista14cid}. These compare two classifiers by comparing the ratio of estimates from the train data with those of the test data to form a kind of contingency table. Computing train estimates through cross validation for TS-CHIEF and InceptionTime is unpractical due to run times. However, it is easy with ROCKET, since it is so fast. Figure~\ref{fig:texas} shows the plot for ROCKET vs HC2. Using the train estimates would lead to the correct decision of choosing HC2 on 94 of the 112 datasets.

\begin{figure}[htb]
    	\centering
        \includegraphics[width=\linewidth,trim={1cm 4cm 0cm 4cm},clip]{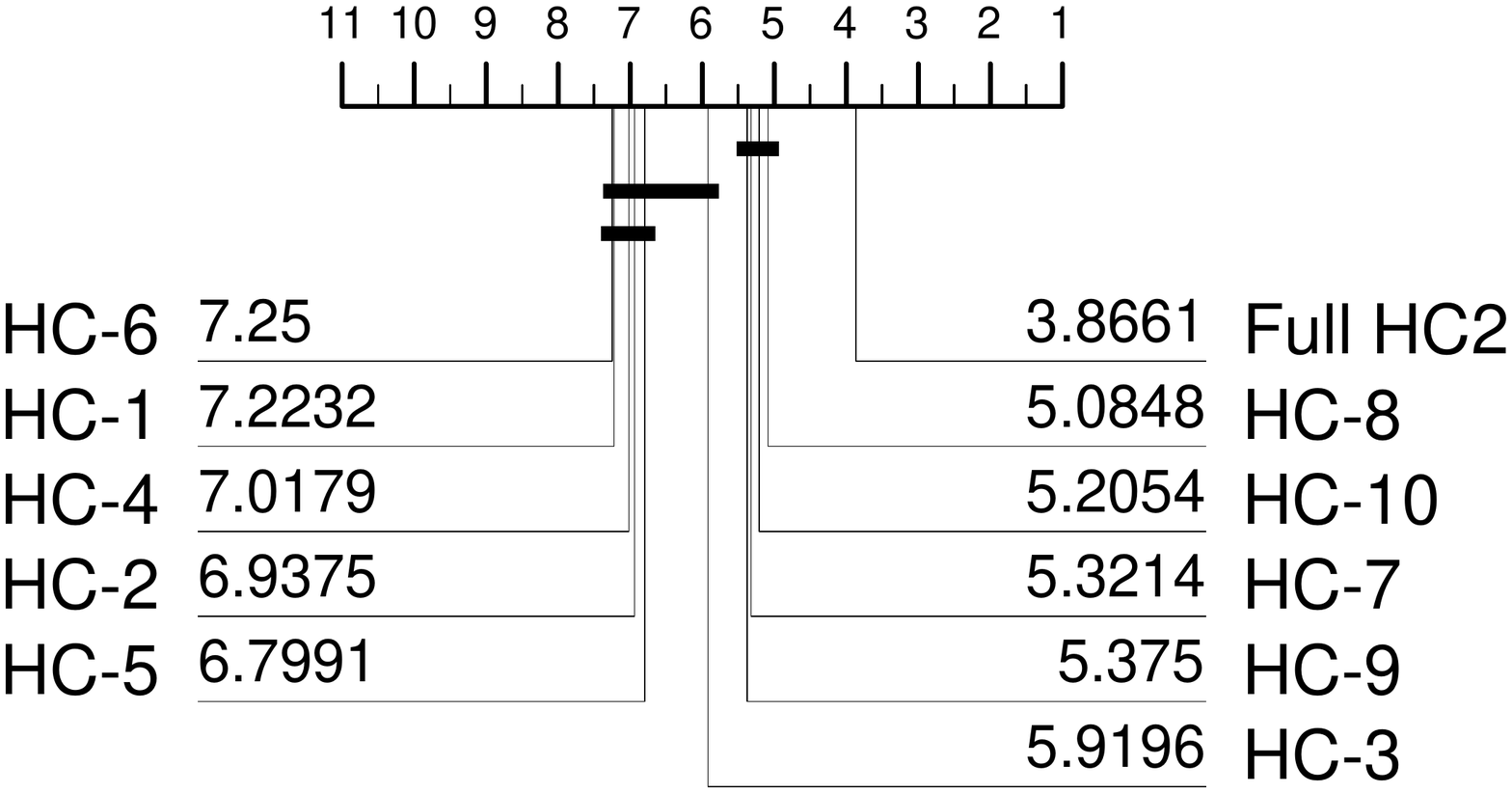}
        \caption{Critical difference diagram for 11 variants of HIVE-COTE 2.0 described in Table~\ref{tab:variants}. Full HC2 contains all four components and is referred to as simply HC2 elsewhere.}
        \label{fig:hc-components}
    \end{figure}

    The next issue is to quantify what impact each component has on the overall performance. Ignoring single component variants (which are presented in Figure~\ref{fig:sota-acc}), there are 11 possible combinations, identified as HC-1 to HC-10 in Table~\ref{tab:variants}, with the eleventh being the Full HC2, referred to as just HC2 elsewhere.
    \begin{table}[htb]

        \centering
        \small
        \begin{tabular}{c|cccccccccc}
Component   & HC-1  &  HC-2  &  HC-3  &   HC-4  &  HC-5  &   HC-6  &  HC-7  &   HC-8  &  HC-9  & HC-10 \\ \hline
DrCIF       &  X    &   X    &  X     &         &        &         &   X    &   X     &   X    &        \\
Arsenal     &  X    &        &        &   X     &   X    &         &   X    &   X     &        &  X       \\
STC         &       &   X    &        &   X     &        &   X     &   X    &         &   X    &  X      \\
TDE         &       &        & X      &         &   X    &   X     &        &   X     &   X    &  X      \\
        \end{tabular}
        \caption{Possible variants of HC2 components.}
        \label{tab:variants}

    \end{table}
Figure~\ref{fig:hc-components} shows the relative performance of the 11 possible variants. The two component models (HC-1 to HC-6) form a clear clique, followed by another clique of three component versions. However, the full four component classifier is significantly more accurate than all of the other variants. This demonstrates that each element contributes to the overall whole.

    %\subsection{HC Module Selection}

    %automate selection of modules per dataset

    \subsection{Ensembling Methods}

    In extending HC1 into HC2, the two main factors are what representations should be included in the ensemble, and how the predictions drawn from each representation should be combined. Here we investigate the latter. HC1 uses the CAWPE ensembling scheme, which was found to be the best combination method for small sets of diverse classifiers across two different dataset archives with limited domain specialisation or prior knowledge~\citep{large19cawpe}. It was also shown that it improved HC1's performance relative to the previous simple majority voting. With updated components, which may be more or less specialised into their own representation formats with different degrees of overlap in their expertise, does this still hold true, or would a different scheme be better? We compare various ensemble selection and stacking schemes to assess whether a more complex scheme than CAWPE could improve HC2. To avoid suspicions of overfitting, we make it clear that we performed this analysis after generating the results presented in Section~\ref{sec:results} using the design we selected a priori.

    For context, we also compare to the individual model selection schemes of picking the best classifier per dataset resample based on the train estimates, and picking the best based on the test data (i.e. cheating) as an oracle scheme. In general, a reasonable ensembling scheme will on average perform somewhere between these two landmarks across arbitrarily large dataset spaces. Combining the predictions of the classifier pool has a beneficial averaging effect by accounting for the imperfections of performance estimation mechanisms from the train data. However, they may have an overly conservative averaging effect compared to looking at test performance to pick the best.

   \begin{figure}[htb]
    	\centering
        \includegraphics[width=\linewidth,trim={3cm 8cm 1cm 4cm},clip]{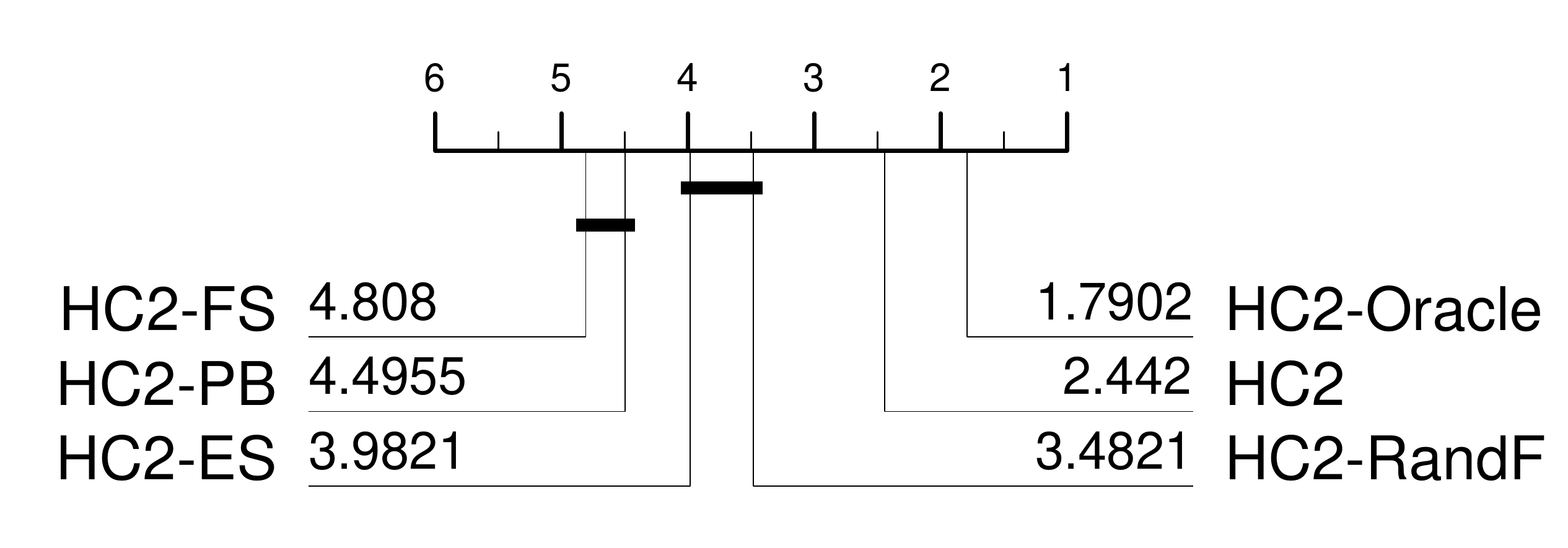}
        \caption{Critical difference diagram comparing different ensemble schemes to default HC2 over the 112 univariate archive datasets.}
        \label{fig:enscompcd}
    \end{figure}

    Figure~\ref{fig:enscompcd} summarises our comparison of alternate ensembling schemes over the HC2 components. HC2-Oracle cheats by picking the single best component based on test accuracy, while HC2-PB picks the best based on the train estimate of test accuracy. HC2-FS executes a forward selection of components per dataset, ranking the train estimates and continuing to include components into the ensemble in order while the ensemble's own train estimate continues to improve. HC2-ES uses ensemble selection per dataset as described in~\citep{caruana04selection}, which selects and weights components based on a repeated bagging with replacement strategy. Lastly, HC2-RandF stacks a random forest classifier onto the meta-data of the components' predicted probability distributions.

\begin{figure}[htb]
    	\centering
        \includegraphics[width=0.8\linewidth,trim={1cm 3cm 1cm 0cm},clip]{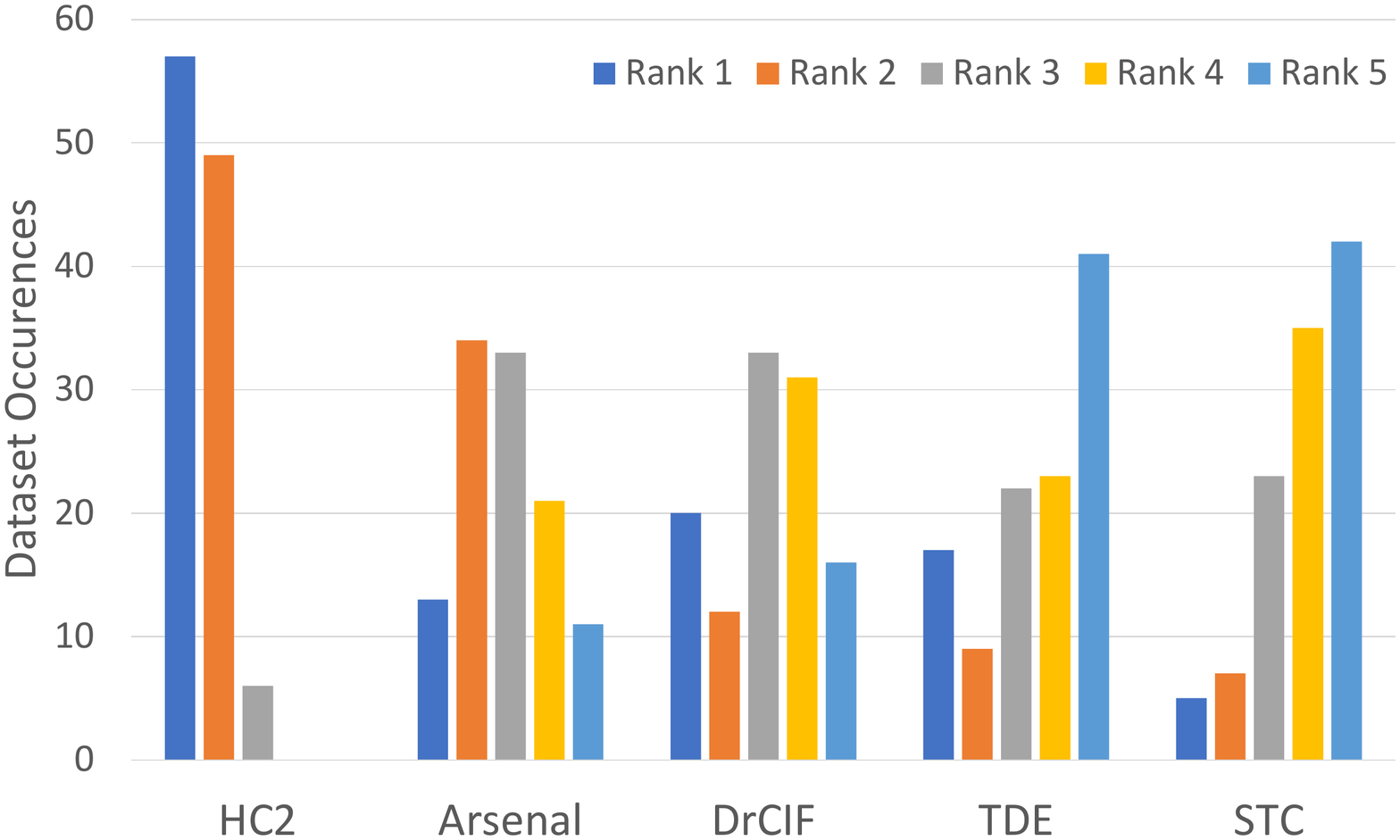}
        \caption{Histograms of ranks between HC2 and its components over the 112 univariate datasets.}
        \label{fig:comprankhist}
    \end{figure}

    We can see that, unsurprising, the oracle selection scheme is still the best on average, and that perfectly selecting the best representation per dataset would still be better than combining them. If there is enough training data to produce reliable train estimates that are unbiased and have very low variance, this may be achievable in practice. However, the reality on our data is that picking the best on the train data (HC-PB) is significantly worse than HC2 and HC2-Oracle. It is worth stressing that selecting the single best component on test accuracy is not always the best. In accordance with the original hypothesis for HIVE-COTE, there are 31 datasets where combining representations is outright better than picking the best, even with perfect hindsight of stochastic differences brought about by resampling.
    %HERE WIN/LOSS vs ORACLE.
    %HC2-Oracle is more accurate on 73 problems than HC2, and less so on 31.
    For many problems, discriminatory features may exist in multiple domains. This is often counter to received wisdom, it is always tempting to think a single type of model is the best approach. HC2 can discover complex interactions between domains. Figure~\ref{fig:comprankhist} compares the ranks of HC2 (with its default CAWPE) and its individual constituents in isolation. HC2 is in fact best or tied for best on 57 of the 112 datasets, and rarely if ever ranked worse than second. This shows that beyond the requirement for perfect domain knowledge being required to beat ensembling on average, on many individual datasets more representations increase the accuracy outright.

    Otherwise, Figure~\ref{fig:enscompcd} shows that using the CAWPE scheme over the HC2 components is significantly better than the alternatives on average. Most popular ensemble schemes in the literature assume a large pool of potentially homogeneous classifiers. We have a small pool of heterogeneous classifiers, and evidence from these experiments and an extensive study on standard classification problems~\citep{large19cawpe} suggest that CAWPE is the best ensemble scheme for this scenario.

\section{HC2 Usability}
\label{sec:usability}
    All our code is open source and our experiments are simple to reproduce. Two implementations of HC2 are available in toolkits we help maintain and develop.  \texttt{tsml}\footnote{https://github.com/uea-machine-learning/tsml} is a Java based time series toolkit compatible with Weka and our primary development platform for TSC. We also implement our algorithms in \texttt{sktime}\footnote{https://github.com/alan-turing-institute/sktime}, a Python based time series toolkit compatible with scikit-learn.
    Where possible, we have verified the consistency of results in both toolkits. Both offer an easy to use interface, and we have provided example code on the website associated with this paper. All datasets are available in a format directly usable in \texttt{tsml} and \texttt{sktime}, and we have also provided details of how to recreate all our experiments.

 %   \subsection{HC2 Contracting}

Table~\ref{tab:runtime} shows that when run sequentially, HC2 is slower than the current state of the art, particularly ROCKET. If speed is more important than a small accuracy gain, this is an argument against using HC2. HC2 is simply not designed to be trained in seconds, and we would not recommend its use in scenarios where models need to trained incredibly quickly. However, we have designed HC2 so that the run time can be controlled by the user through a time contract. We make the assumption that when time is a serious constraint, the problem must be fairly large. There are only five problems where a sequential build of HC2 would take more than half a day with a single processor run. The list of problems, and the time taken by TS-CHIEF, InceptionTime and ROCKET, are listed in Table~\ref{tab:slow}. ROCKET is very fast, InceptionTime hard to compare to and TS-CHIEF is similar to HC2 on average but is unpredictable.
\begin{table}[htb]
    \centering
    \caption{Train times in hours on problems where a sequential run of HC2 takes longer than 12 hours.}
    \label{tab:slow}
    \begin{tabular}{c|c|c|c|c|c}
    Problem     & Rocket    & InceptionTime & TS-CHIEF & Hc1 & HC2 \\ \hline
ElectricDevices & 0.43 & 	6.15 & 	7.94   & 	56.80  & 	54.55\\
Crop            & 0.22 & 	4.37 & 	4.59    & 	73.54 & 	28.41\\
FordB           & 0.19 & 	6.06 & 	43.33   & 	19.33 & 	24.67 \\
FordA           & 0.19 & 	6.01 & 	40.88	 & 20.25 & 	23.52\\
HandOutlines    & 0.23 & 	7.11 & 	166.76	 & 7.77	 & 18.53\\
    \end{tabular}
\end{table}
If we run HC2 with a four hour contract on these problems we achieve 98\% of the final accuracy, and if we run it for 12 hours we achieve 99\% of the full build accuracy. If a reasonable model on bigger problems is required in hours, then contracting HC2 offers a good solution. However, if the problem is truly large, then all TSC have usability issues. ROCKET, TS-CHIEF can require massive amounts of memory and/or time.
%    \subsection{HC2 for Large Problems}
%    \label{subsec: Large problems}

    %State of play...
    %(Got a load of stuff for ROCKET, CIF etc from before the components changed.)
    %TDE:
    %   Mos - producing 1/2 checkpoints per 7 days! Running some test atm.  (01/04/21). Timed out trying to get 3rd checkpoint (7 day queue).
    %STC: What we have is the limit. Timing out! But have data for all 3, but acc is still climbing at limit.
    %DrCIF:
    %   Ins - Produces 0 checkpoints in 7 days. Probs due to initial transform. (01/04/21)
    %   Mos - Produces 0 checkpoints in 7 days. Probs due to initial transform. (01/04/21)
    %Arsenal:
    %   Ins - Got a number of checkpoints produced, currently churning through test. (01/04/21)
    %   Mos - No checkpoints, having issues with mem, now at 700Gb on hmem but stuck pending behind Arsenal/Ins stuff. (01/04/21)

   For genuinely large data where HC2 may takes weeks for a full run, it is worthwhile considering how long it would take to converge. With many algorithms, it is often the case that most of the gain in accuracy is made relatively quickly. On large datasets this can equate to days of processing that contributes relatively little to the overall performance. Furthermore, a practitioners concern may not be accuracy and an understanding of the evolution of performance over time provides a foundation on which paramerterisation decisions can be made.

    \begin{table}[htb]
        \centering
		\caption{Table showing the attributes of 2 the large datasets used throughout section \ref{subsec: Large problems}.}
		\begin{tabular}{llll}
			\hline
			& \multicolumn{1}{c}{FruitFlies} & \multicolumn{1}{c}{InsectSound} \\ \hline \hline
            Train Size      & 17,255    & 25,000    \\
            Test Size       & 17,256    & 25,000    \\
            No. Classes     & 3         & 10        \\
            Series Length   & 5,000     & 15,883    \\
            Total size (Gb) & 2.44      & 14.75     \\
            \hline
		\end{tabular}
		\label{table:largedata}
	\end{table}
	
    In this Section we comment on the evolution of accuracy over time for each of the HC2 components. Experiments were run on 2 large datasets, described in table~\ref{table:largedata}. The datasets used are notoriously problematic for complex approaches. This is typically because internal transforms are sensitive to series length, number of cases or both. These datasets were deliberately chosen to explore the limitations of the HC2 constituents.

    \begin{table}[htb]
        \centering
        \caption{Table showing both accuracy achieved by last checkpoint and variance in accuracy between first and last checkpoint for each HC2 constituent on the FruitFlies dataset.}
        \begin{tabular}{llll}
            \hline
                    & Accuracy & Variance & Constituents \\
            \hline \hline
            TDE     & 0.8016   & -0.0822  & 20           \\
            STC     & 0.8901   & -0.0431  & N/A          \\
            DrCIF   & 0.9278   & -0.1084  & 65           \\
            Arsenal & 0.6433   & -0.0131  & 25           \\
            \hline
        \end{tabular}
        \label{table:FuitFlies_CP}
    \end{table}

\begin{figure}[htb]
    	\centering
        \includegraphics[width=\linewidth,trim={1cm 10cm 2cm 10cm},clip]{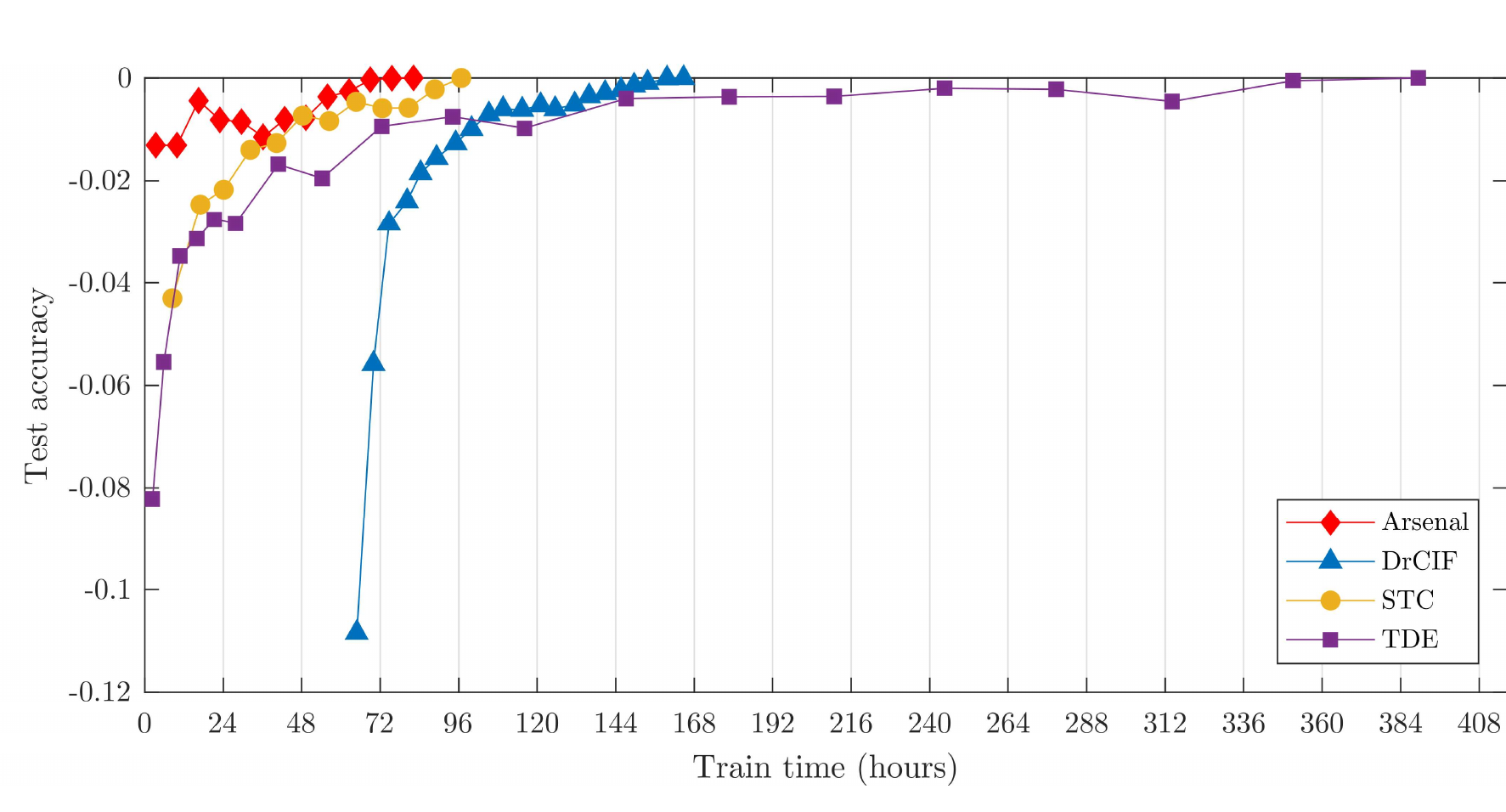}
        \caption{Accuracy as a function of train time for HC2 components on the FruitFlies dataset.}
        \label{fig:Checkpoint FruitFlies}
    \end{figure}

\begin{figure}[htb]
    	\centering
        \includegraphics[width=\linewidth,trim={1cm 10cm 2cm 10cm},clip]{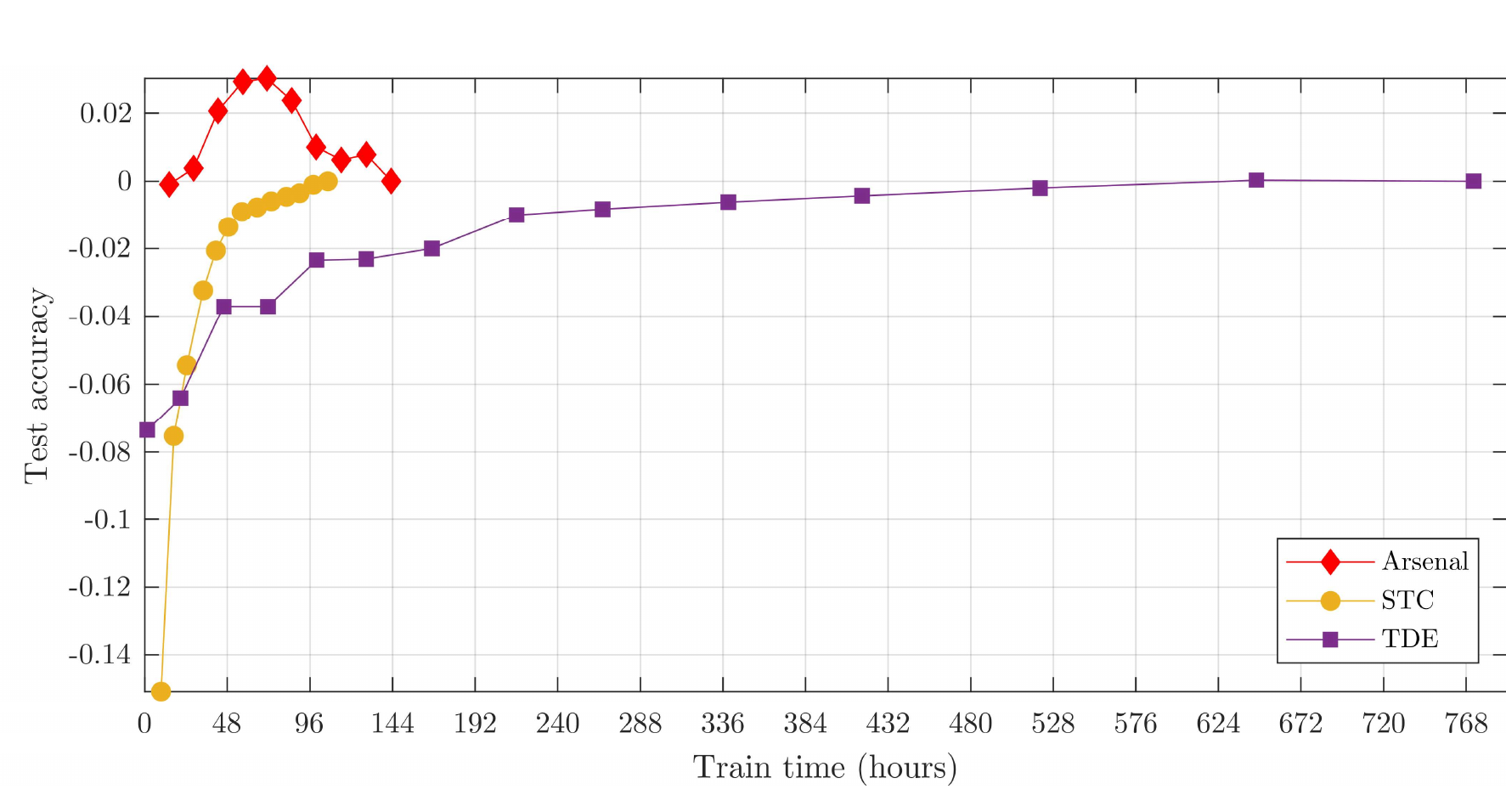}
        \caption{Accuracy as a function of train time for HC2 components on the InsectWingbeat dataset.}
        \label{fig:Checkpoint Insect}
    \end{figure}

    In order to overcome the imposed runtime limitations on the UEA HPC a checkpointing mechanism was utilised to periodically save an approaches state during the training phase. This allowed both the continuation of training beyond what would usually be feasible, via reloading and continuing training from the saved point, and the opportunity to assess the approach at each saved state, via invoking the test phase after reloading the saved state. Figures~\ref{fig:Checkpoint FruitFlies}~\&~\ref{fig:Checkpoint Insect} show how relative accuracy changes with respect to time for the FruitFlies and InsectSound datasets. Each data point shows the relative difference of the accuracy achieved at each checkpoint with respect to the last checkpoint recorded.
    %(CLARIFY IS IT THE "FINAL" ACCURACY? WHY IS ARSENAL POSITIVE?).
    Tables~\ref{table:FuitFlies_CP}~\&~\ref{table:Insect_CP} present the real accuracy achieved with the last checkpoint processed, the variance in accuracy between the first and last recorded checkpoint, and the number of constituents built in by the last checkpoint.

    \begin{table}[htb]
        \centering
        \caption{Table showing both accuracy achieved by last checkpoint and variance in accuracy between first and last checkpoint for each HC2 constituent on the InsectSound dataset.}
        \begin{tabular}{llll}
        \hline
                & Accuracy & Variance & Constituents \\
        \hline \hline
        TDE     & 0.2746   & -0.0735  & 14           \\
        STC     & 0.7357   & -0.1509  & N/A          \\
        DrCIF   & -        & -        & -            \\
        Arsenal & 0.2951   & -0.0313  & 10           \\
        \hline
        \end{tabular}
        \label{table:Insect_CP}
    \end{table}

    Figures~\ref{fig:Checkpoint FruitFlies}~\&~\ref{fig:Checkpoint Insect} show the accuracy of DrCIF, STC and TDE follow a similar trend with respect to time. In the case of these approaches accuracy increases quickly as the number of constituents present increases, before reaching the point of diminishing returns. For these approaches 80\% of the accuracy achieved is done so in less than 50\% of the train time used by the last recorded checkpoint. The Arsenal approach does not follow this trend and instead the total variance in accuracy throughout the training period is not as pronounced. Also, the changes in accuracy appear to be more erratic with additional constituents producing decreases in accuracy as well as increases.

    Furthermore, of the 8 combinations presented only the Arsenal approach on the FruitFlies dataset was able to complete. In most cases the outstanding experiments were prohibited by inflation in the time taken to test. Checkpointing during the test phase is not implemented and as a result the entire test process is subject to a hard time limit of 7 days. This effected STC and DrCIF on the FruitFlies dataset and STC and TDE on the InsectSounds dataset. Aditionally, the Arsenal approach was limited by its memory requirement on the InsectSounds dataset, for which we are limited to 700Gb.

\section{Conclusion}
\label{sec:conclusion}

HIVE-COTE version 2.0 is a meta ensemble of four very different classifiers, each of which is designed to capture different discriminatory features.
It represents a new state of the art in terms of time series classification, significantly outperforming the previous best on both univariate and multivariate problems in terms of accuracy. Our ablative study showed that HC2 is better than any one of its constituents, and that each component makes a significant contribution to the overall performance. We believe its strength lies in the fact that many problems have discriminatory features in multiple data domains; a shapelet might be indicative of one class value, whereas a repeating pattern may characterise another. HC2 uses a simple yet highly effective ensemble scheme to combine this information which we demonstrated was significantly better than alternatives such as stacking or a selection strategy.

HC2's weakness is that it does not scale well to very large problems. We showed that for problems with thousands of series of length in the tens of thousands, build times can be excessive, but that contacting means you can at least get an estimate. We note that the current state of the art has similar limitations. Even ROCKET, which is by far the fastest algorithm, does not scale well with increasing number of instances.

There is room for further improvement with HC2. The STC design remains the same in HC1 and HC2, and we believe there are ways it could be improved. Contracting could be enhanced so that components could produce bespoke estimates of the time, and reconfigure themselves more intelligently to make best use of the available run time. Individual components could be threaded. Variability in estimates from the train data could be incorporated into ensemble process, and weights per instance might improve predictions.

HC2 is available in two open source toolkits and has improved usability features such as contracting, which allow the user to specify an approximate maximum run time. Our experiments are easily reproducible, and an accompanying website contains complete results and more information on how to use HC2.

\section*{Acknowledgements}{
    This work is supported by the UK Engineering and Physical Sciences Research Council (EPSRC) iCASE award T206188 sponsored by British Telecom. The experiments were carried out on the High Performance Computing Cluster supported by the Research and Specialist Computing Support service at the University of East Anglia.
}

%supplementary stuff:

%cv vs oob


\begin{thebibliography}{42}
\providecommand{\natexlab}[1]{#1}
\providecommand{\url}[1]{{#1}}
\providecommand{\urlprefix}{URL }
\expandafter\ifx\csname urlstyle\endcsname\relax
  \providecommand{\doi}[1]{DOI~\discretionary{}{}{}#1}\else
  \providecommand{\doi}{DOI~\discretionary{}{}{}\begingroup
  \urlstyle{rm}\Url}\fi
\providecommand{\eprint}[2][]{\url{#2}}

\bibitem[{Arul and Kareem(2021)}]{arul2021applications}
Arul M, Kareem A (2021) Applications of shapelet transform to time series
  classification of earthquake, wind and wave data. Engineering Structures
  228:111564

\bibitem[{Bagnall et~al.(2017)Bagnall, Lines, Bostrom, Large, and
  Keogh}]{bagnall17bakeoff}
Bagnall A, Lines J, Bostrom A, Large J, Keogh E (2017) The great time series
  classification bake off: a review and experimental evaluation of recent
  algorithmic advances. Data Mining and Knowledge Discovery 31(3):606--660

\bibitem[{Bagnall et~al.(2018)Bagnall, Dau, Lines, Flynn, Large, Bostrom,
  Southam, and Keogh}]{bagnall18mtsc}
Bagnall A, Dau H, Lines J, Flynn M, Large J, Bostrom A, Southam P, Keogh E
  (2018) The {UEA} multivariate time series classification archive, 2018. ArXiv
  e-prints arXiv:1811.00075, \urlprefix\url{http://arxiv.org/abs/1809.06705}

\bibitem[{Bagnall et~al.(2020)Bagnall, Flynn, Large, Lines, and
  Middlehurst}]{bagnall20hivecote1}
Bagnall A, Flynn M, Large J, Lines J, Middlehurst M (2020) On the usage and
  performance of {HIVE-COTE v1.0}. In: proceedings of the 5th Workshop on
  Advances Analytics and Learning on Temporal Data, Lecture Notes in Artificial
  Intelligence, vol 12588

\bibitem[{Batista et~al.(2014)Batista, Keogh, Tataw, and
  deSouza}]{batista14cid}
Batista G, Keogh E, Tataw O, deSouza V (2014) {CID}: an efficient
  complexity-invariant distance measure for time series. Data Mining and
  Knowledge Discovery 28(3):634--669

\bibitem[{Benavoli et~al.(2016)Benavoli, Corani, and
  Mangili}]{benavoli16pairwise}
Benavoli A, Corani G, Mangili F (2016) Should we really use post-hoc tests
  based on mean-ranks? Journal of Machine Learning Research 17:1--10

\bibitem[{Bostrom and Bagnall(2017)}]{bostrom17binary}
Bostrom A, Bagnall A (2017) Binary shapelet transform for multiclass time
  series classification. Transactions on Large-Scale Data and Knowledge
  Centered Systems 32:24--46

\bibitem[{Cabello et~al.(2020)Cabello, Naghizade, Qi, and
  Kulik}]{cabello20fast}
Cabello N, Naghizade E, Qi J, Kulik L (2020) Fast and accurate time series
  classification through supervised interval search. In: proceedings of the
  {IEEE} International Conference on Data Mining

\bibitem[{Caruana and Niculescu-Mizil(2004)}]{caruana04selection}
Caruana R, Niculescu-Mizil A (2004) Ensemble selection from libraries of
  models. In: Proc. of the 21st International Conference on Machine learning

\bibitem[{Chaovalitwongse et~al.(2006)Chaovalitwongse, Prokopyev, and
  Pardalos}]{chaovalitwongse2006electroencephalogram}
Chaovalitwongse WA, Prokopyev OA, Pardalos PM (2006) Electroencephalogram (eeg)
  time series classification: Applications in epilepsy. Annals of Operations
  Research 148(1):227--250

\bibitem[{Dau et~al.(2019)Dau, Bagnall, Kamgar, Yeh, Zhu, Gharghabi,
  Ratanamahatana, Chotirat, and Keogh}]{dau19ucr}
Dau H, Bagnall A, Kamgar K, Yeh M, Zhu Y, Gharghabi S, Ratanamahatana C,
  Chotirat A, Keogh E (2019) The {UCR} time series archive. IEEE/CAA Journal of
  Automatica Sinica 6(6):1293--1305

\bibitem[{Dempster et~al.(2020)Dempster, Petitjean, and
  Webb}]{dempster20rocket}
Dempster A, Petitjean F, Webb G (2020) {ROCKET}: Exceptionally fast and
  accurate time series classification using random convolutional kernels. Data
  Mining and Knowledge Discovery 34:1454--1495

\bibitem[{Dem\v{s}ar(2006)}]{demsar06comparisons}
Dem\v{s}ar J (2006) Statistical comparisons of classifiers over multiple data
  sets 7:1--30

\bibitem[{Deng et~al.(2013)Deng, Runger, Tuv, and Vladimir}]{deng13forest}
Deng H, Runger G, Tuv E, Vladimir M (2013) A time series forest for
  classification and feature extraction. Information Sciences 239:142--153

\bibitem[{Fawaz et~al.(2019)Fawaz, Forestier, Weber, Idoumghar, and
  Muller}]{fawaz19deep}
Fawaz H, Forestier G, Weber J, Idoumghar L, Muller P (2019) Deep learning for
  time series classification: a review. Data Mining and Knowledge Discovery
  33(4):917--963

\bibitem[{Fawaz et~al.(2020)Fawaz, Lucas, Forestier, Pelletier, Schmidt, Weber,
  Webb, Idoumghar, Muller, and Petitjean}]{fawaz20inception}
Fawaz H, Lucas B, Forestier G, Pelletier C, Schmidt D, Weber J, Webb G,
  Idoumghar L, Muller P, Petitjean F (2020) {InceptionTime}: finding {AlexNet}
  for time series classification. Data Mining and Knowledge Discovery
  34(6):1936--1962

\bibitem[{Fulcher and Jones(2017)}]{fulcher17hctsa}
Fulcher B, Jones N (2017) hctsa: A computational framework for automated
  time-series phenotyping using massive feature extraction. Cell Systems
  5(5):527--531

\bibitem[{Garc\'{i}a and Herrera(2008)}]{garcia08pairwise}
Garc\'{i}a S, Herrera F (2008) An extension on “statistical comparisons of
  classifiers over multiple data sets” for all pairwise comparisons
  9:2677--2694

\bibitem[{Guillaume et~al.(2020)Guillaume, Vrain, and Wael}]{guillaume2020time}
Guillaume A, Vrain C, Wael E (2020) Time series classification for predictive
  maintenance on event logs. arXiv preprint arXiv:201110996

\bibitem[{Hills et~al.(2014)Hills, Lines, Baranauskas, Mapp, and
  Bagnall}]{hills14shapelet}
Hills J, Lines J, Baranauskas E, Mapp J, Bagnall A (2014) Classification of
  time series by shapelet transformation. Data Mining and Knowledge Discovery
  28(4):851--881

\bibitem[{Large et~al.(2019{\natexlab{a}})Large, Bagnall, Malinowski, and
  Tavenard}]{large19dictionary}
Large J, Bagnall A, Malinowski S, Tavenard R (2019{\natexlab{a}}) On time
  series classification with dictionary-based classifiers. Intelligent Data
  Analysis 23(5)

\bibitem[{Large et~al.(2019{\natexlab{b}})Large, Lines, and
  Bagnall}]{large19cawpe}
Large J, Lines J, Bagnall A (2019{\natexlab{b}}) A probabilistic classifier
  ensemble weighting scheme based on cross validated accuracy estimates. Data
  Mining and Knowledge Discovery 33(6):1674--–1709

\bibitem[{Lazebnik et~al.(2006)Lazebnik, Schmid, and Ponce}]{lazebnik06pyramid}
Lazebnik S, Schmid C, Ponce J (2006) Beyond bags of features: Spatial pyramid
  matching for recognizing natural scene categories. In: 2006 IEEE Computer
  Society Conference on Computer Vision and Pattern Recognition (CVPR'06),
  IEEE, vol~2, pp 2169--2178

\bibitem[{Lines and Bagnall(2015)}]{lines15elastic}
Lines J, Bagnall A (2015) Time series classification with ensembles of elastic
  distance measures. Data Mining and Knowledge Discovery 29:565--592

\bibitem[{Lines et~al.(2016)Lines, Taylor, and Bagnall}]{lines16hive}
Lines J, Taylor S, Bagnall A (2016) {HIVE-COTE}: The hierarchical vote
  collective of transformation-based ensembles for time series classification.
  In: proceedings of 16th {IEEE} International Conference on Data Mining

\bibitem[{Lines et~al.(2018)Lines, Taylor, and Bagnall}]{lines18hive}
Lines J, Taylor S, Bagnall A (2018) Time series classification with
  {HIVE-COTE}: The hierarchical vote collective of transformation-based
  ensembles. ACM Transactions Knowledge Discovery from Data 12(5):1--36

\bibitem[{Lubba et~al.(2019)Lubba, Sethi, Knaute, Schultz, Fulcher, and
  Jones}]{lubba19catch22}
Lubba C, Sethi S, Knaute P, Schultz S, Fulcher B, Jones N (2019) catch22:
  canonical time-series characteristics. Data Mining and Knowledge Discovery
  33(6):1821--1852

\bibitem[{Lucas et~al.(2019)Lucas, Shifaz, Pelletier, O’Neill, Zaidi,
  Goethals, Petitjean, and Webb}]{lucas19proximity}
Lucas B, Shifaz A, Pelletier C, O’Neill L, Zaidi N, Goethals B, Petitjean F,
  Webb G (2019) Proximity forest: an effective and scalable distance-based
  classifier for time series. Data Mining and Knowledge Discovery
  33(3):607--635

\bibitem[{Middlehurst et~al.(2019)Middlehurst, Vickers, and
  Bagnall}]{middlehurst19scalable}
Middlehurst M, Vickers W, Bagnall A (2019) Scalable dictionary classifiers for
  time series classification. In: proceedings of Intelligent Data Engineering
  and Automated Learning, Lecture Notes in Computer Science, vol 11871, pp
  11--19

\bibitem[{Middlehurst et~al.(2020{\natexlab{a}})Middlehurst, Large, and
  Bagnall}]{middlehurst20canonical}
Middlehurst M, Large J, Bagnall A (2020{\natexlab{a}}) The canonical interval
  forest {(CIF)} classifier for time series classification. In: proceedings of
  the {IEEE} International Conference on Big Data

\bibitem[{Middlehurst et~al.(2020{\natexlab{b}})Middlehurst, Large, Cawley, and
  Bagnall}]{middlehurst20temporal}
Middlehurst M, Large J, Cawley G, Bagnall A (2020{\natexlab{b}}) The temporal
  dictionary ensemble {(TDE)} classifier for time series classification. In:
  proceedings of the European Conference on Machine Learning and Principles and
  Practice of Knowledge Discovery in Databases

\bibitem[{Nguyen et~al.(2019)Nguyen, Gsponer, Ilie, O'Reilly, and
  Ifrim}]{nguyen19interpretable}
Nguyen TL, Gsponer S, Ilie I, O'Reilly M, Ifrim G (2019) Interpretable time
  series classification using linear models and multi-resolution multi-domain
  symbolic representations. Data Mining and Knowledge Discovery
  33(4):1183--1222

\bibitem[{Potamitis(2014)}]{potamitis2014classifying}
Potamitis I (2014) Classifying insects on the fly. Ecological informatics
  21:40--49

\bibitem[{Rodriguez et~al.(2006)Rodriguez, Kuncheva, and
  Alonso}]{rodriguez06rotf}
Rodriguez J, Kuncheva L, Alonso C (2006) Rotation forest: A new classifier
  ensemble method. {IEEE} Transactions on Pattern Analysis and Machine
  Intelligence 28(10):1619--1630

\bibitem[{Ruiz et~al.(2021)Ruiz, Flynn, Large, Middlehurst, and
  Bagnall}]{ruiz21mtsc}
Ruiz AP, Flynn M, Large J, Middlehurst M, Bagnall A (2021) The great
  multivariate time series classification bake off: a review and experimental
  evaluation of recent algorithmic advances. Data Mining and Knowledge
  Discovery 35(2):401--–449

\bibitem[{Sch{\"a}fer(2015)}]{schafer15boss}
Sch{\"a}fer P (2015) The {BOSS} is concerned with time series classification in
  the presence of noise. Data Mining and Knowledge Discovery 29(6):1505--1530

\bibitem[{Sch{\"a}fer and H{\"o}gqvist(2012)}]{schafer12sfa}
Sch{\"a}fer P, H{\"o}gqvist M (2012) {SFA: a symbolic {Fourier} approximation
  and index for similarity search in high dimensional datasets}. In:
  proceedings of the 15th International Conference on Extending Database
  Technology, pp 516--527

\bibitem[{Sch{\"a}fer and Leser(2017{\natexlab{a}})}]{schafer17fast}
Sch{\"a}fer P, Leser U (2017{\natexlab{a}}) Fast and accurate time series
  classification with {WEASEL}. In: proceedings of the ACM on Conference on
  Information and Knowledge Management, pp 637--646

\bibitem[{Sch{\"a}fer and Leser(2017{\natexlab{b}})}]{schafer17multivariate}
Sch{\"a}fer P, Leser U (2017{\natexlab{b}}) Multivariate time series
  classification with weasel+ muse. arXiv preprint arXiv:171111343

\bibitem[{Shifaz et~al.(2020)Shifaz, Pelletier, Petitjean, and
  Webb}]{shifaz20ts-chief}
Shifaz A, Pelletier C, Petitjean F, Webb G (2020) {TS-CHIEF}: A scalable and
  accurate forest algorithm for time series classification. Data Mining and
  Knowledge Discovery pp 1--34

\bibitem[{Shokoohi-Yekta et~al.(2017)Shokoohi-Yekta, Hu, Jin, Wang, and
  Keogh}]{shokoohi2017generalizing}
Shokoohi-Yekta M, Hu B, Jin H, Wang J, Keogh E (2017) Generalizing dtw to the
  multi-dimensional case requires an adaptive approach. Data mining and
  knowledge discovery 31(1):1--31

\bibitem[{Szegedy et~al.(2015)Szegedy, Liu, Jia, Sermanet, Reed, Anguelov,
  Erhan, Vanhoucke, and Rabinovich}]{szegedy15inception}
Szegedy C, Liu W, Jia Y, Sermanet P, Reed S, Anguelov D, Erhan D, Vanhoucke V,
  Rabinovich A (2015) Going deeper with convolutions. In: The IEEE Conference
  on Computer Vision and Pattern Recognition (CVPR)

\end{thebibliography}
\end{document}